\documentclass{article} %
\usepackage{iclr2027_conference,times}

\usepackage{amsmath,amsfonts,bm}

\def\eqref#1{equation~\ref{#1}}

\def\1{\bm{1}}

\DeclareMathAlphabet{\mathsfit}{\encodingdefault}{\sfdefault}{m}{sl}
\SetMathAlphabet{\mathsfit}{bold}{\encodingdefault}{\sfdefault}{bx}{n}

\usepackage{xcolor}
\definecolor{linkc}{rgb}{0, 0.44, 0.74}
\definecolor{eqc}{rgb}{0, 0.5, 0}
\usepackage[pagebackref=false,breaklinks=true,colorlinks=true,citecolor=linkc,linkcolor=eqc,bookmarks=false]{hyperref}
\usepackage{url}

\usepackage{multirow}
\usepackage{adjustbox}
\usepackage{booktabs}
\usepackage{enumitem}
\usepackage{array}
\usepackage{colortbl}
\usepackage{subcaption}
\usepackage{siunitx}      %
\usepackage{makecell}     %
\usepackage{threeparttable} %
\usepackage{tcolorbox}
\tcbuselibrary{breakable}
\usepackage{graphicx}
\usepackage{csquotes}
\usepackage{tabularx}
\usepackage{tabularray}
\usepackage{wrapfig}  %
\usepackage{float}
\usepackage[normalem]{ulem}     %
\usepackage{pifont}

\newcommand{\OURS}{VideoEvolve: Co-Evolving Memory and Retrieval for Long Video Understanding}

\title{\OURS}

\author{
Yongchao Xu$^{1,2}$,
Bowen Ye$^{3}$,
Jiefeng Gan$^{1}$,
Junkai Ma$^{1}$\thanks{Corresponding authors.},
Wenzhao Li$^{1}$\thanks{Project leader.}, 
 Sen Tao$^{4}$,\\
\bfseries Yi Wei$^{5}$,
Jiawei Liu$^{2}$\footnotemark[1]
\\
$^{1}$Alibaba Group, Hangzhou, China \\
$^{2}$University of Science and Technology of China, Hefei, China \\
$^{3}$Shanghai Jiao Tong University, Shanghai, China \\
$^{4}$University of Chinese Academy of Sciences, Beijing, China \\
$^{5}$Beihang University, Beijing, China \\
\\
}

\iclrfinalcopy %
\begin{document}

\maketitle
\lhead{Preprint}

\begin{abstract}

Long video understanding increasingly relies on external memory to organize massive visual streams into compact representations. 
However, most memory-based methods dynamically adapt how information is retrieved for different questions, while largely fixing what is remembered. 
This mismatch makes missing details costly to recover, whereas stored information is valuable only when it can be reliably retrieved.
To address this issue, we propose VideoEvolve, a novel self-evolving framework that jointly evolves memory and retrieval for long video understanding. 
Specifically, starting from a coarse low-frame-rate overview, VideoEvolve couples a Memory Evolver for selective memory augmentation with a Retrieval Evolver for adaptive retrieval over the evolving memory. 
We then co-evolve the two Evolvers through alternating agentic reinforcement learning (Agentic RL), updating one while freezing the other.
To steer this alternating evolution, Bottleneck-Aware Evolution Feedback (BEF) identifies whether the current bottleneck lies in memory or retrieval and directs optimization toward the more limiting side.
Furthermore, VideoEvolve introduces Capability-Aware Evolution Feedback (CEF) to alleviate downstream feedback from over-specializing memory to a fixed set of training questions, shifting training toward underdeveloped yet learnable video capabilities.
By integrating Agentic RL with BEF and CEF, VideoEvolve transforms downstream reasoning experience into transferable capability updates, providing a concrete path from static long-video systems toward experience-driven, self-improving multimodal intelligence. 
Extensive experiments on multiple long video understanding benchmarks demonstrate the effectiveness of VideoEvolve.

\end{abstract}

\section{Introduction}

Long video understanding remains challenging for Multimodal Large Language Models (MLLMs), as task-relevant evidence is often sparsely distributed within videos that can last over an hour, while exhaustive visual processing incurs substantial computational costs~\citep{wu2024longvideobench,fu2025video,wang2025lvbench}.
To address this challenge, recent agentic Video-MLLMs adopt two representative paradigms for selective information access: \textit{Video-Revisiting Methods}, which localize and revisit question-relevant video segments to recover fine-grained visual evidence~\citep{ding2026videozoomer,yang2026longvt}; and \textit{Memory-Based Methods}, which organize video content into persistent memory and adaptively retrieve relevant evidence from this compact representation~\citep{song2024moviechat,yeo2026worldmm}.
By reusing stored information across questions, memory-based methods offer a scalable approach to long video understanding, with the quality of memory construction shaping the evidence available for subsequent retrieval and reasoning.

\begin{figure*}[t]
    \centering
    \includegraphics[width=\textwidth]{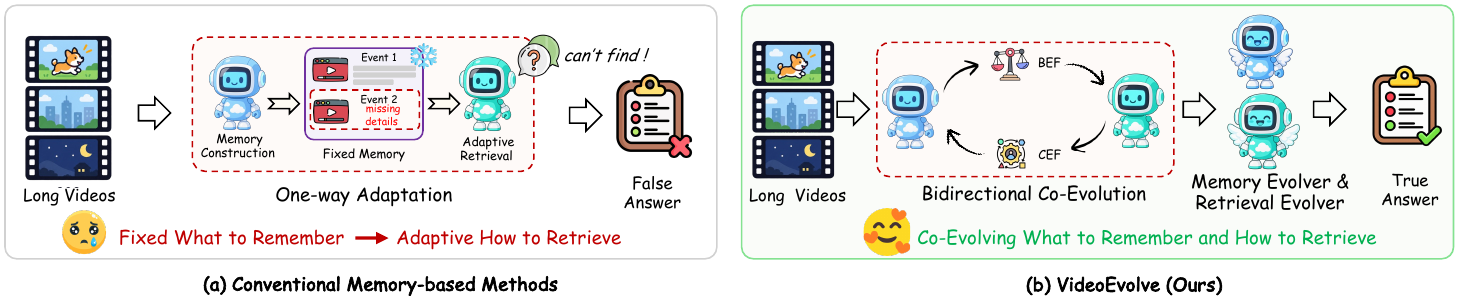}
    \vspace{-18pt}
    \caption{\textbf{Comparison between conventional memory-based methods and VideoEvolve.}
    Conventional methods largely rely on predefined memory construction and decoupled construction–retrieval optimization. VideoEvolve instead co-evolves what to remember and how to retrieve through alternating Agentic RL and downstream reasoning feedback. BEF targets the more limiting component, while CEF prioritizes underdeveloped yet learnable capabilities.
    }
    \vspace{-10pt}
    \label{fig:intro}
\end{figure*}

Despite these advantages, existing memory-based methods largely follow a one-way adaptation paradigm (Figure~\ref{fig:intro}(a)), with two related limitations.
\textbf{(1) Limited Memory Adaptability}: predefined construction policies rarely learn from downstream reasoning feedback~\citep{yeo2026worldmm,choi2026keep}, so omitted evidence can repeatedly force costly raw-video revisits.
\textbf{(2) Memory--Retrieval Misalignment}: construction and retrieval are typically optimized separately~\citep{yang2025egolife,lin2026hippomm}, without explicitly aligning what memory preserves with what retrieval can effectively access.
This raises our central question: can memory construction and retrieval learn to improve each other, enabling \textit{co-evolution} of what to remember and how to retrieve?

To answer this question, we propose \textbf{VideoEvolve} (Figure~\ref{fig:intro}(b)), a self-evolving framework that co-evolves memory construction and retrieval policies through alternating \textbf{agentic reinforcement learning (Agentic RL)}.
Starting from a reusable memory built from low-frame-rate video observations, the Memory Evolver learns to selectively augment stored evidence, while the Retrieval Evolver learns to search the evolving memory and revisit the raw video when needed.
With the Retrieval Evolver frozen, downstream reasoning feedback guides the Memory Evolver to improve memory construction; the Retrieval Evolver is then optimized over the updated, fixed memory.
This alternating process aligns what the system remembers with how it retrieves.
To guide subsequent cycles, \textbf{Bottleneck-Aware Evolution Feedback (BEF)} compares memory-only and revisit-enabled reasoning, accounting for actual video revisits, to estimate the relative training needs of the two Evolvers and direct optimization toward the more limiting side.

While BEF allocates training effort according to the current memory--retrieval bottleneck, this allocation alone does not ensure the broad capability development required for generalizable long video understanding~\citep{fu2026video}.
Repeated optimization on a fixed set of training questions may instead encourage question-specific specialization, motivating capability-level guidance for the co-evolution process.
To address this issue, we introduce \textbf{Capability-Aware Evolution Feedback (CEF)}, which translates capability-level feedback into an adaptive training curriculum.
Specifically, CEF reweights training toward underdeveloped yet learnable capabilities while preserving broad coverage, and adjusts the sampling distribution as the two Evolvers improve.
By integrating bottleneck-aware resource allocation with capability-aware curriculum adaptation, VideoEvolve supports targeted and transferable co-evolution of memory and retrieval.

Our contributions are summarized as follows:
(1) We propose VideoEvolve, a self-evolving framework that co-evolves memory construction and retrieval through alternating Agentic RL, enabling both policies to improve from downstream reasoning feedback.
(2) We introduce Bottleneck-Aware Evolution Feedback (BEF) and Capability-Aware Evolution Feedback (CEF), which respectively adapt training allocation to memory--retrieval bottlenecks and adjust the training curriculum toward underdeveloped yet learnable capabilities.
(3) Experiments on Video-MME~\citep{fu2025video},
LongVideoBench~\citep{wu2024longvideobench}, LVBench~\citep{wang2025lvbench},
MLVU~\citep{zhou2025mlvu}, and MMVU~\citep{zhao2025mmvu} show that VideoEvolve
consistently outperforms the evaluated state-of-the-art open-source baselines.

\section{Related Works}
\textbf{Long Video Understanding with MLLMs.}
Recent MLLMs have extended video understanding toward longer temporal contexts through time-aware modeling, long-context scaling, and adaptive visual compression~\citep{ren2024timechat,shen2024longvu,chen2025longvila}. Beyond directly processing more visual tokens, another line of work shifts toward selective and agentic video access. VideoAgent~\citep{wang2024videoagent} iteratively retrieves task-relevant visual information, DVD~\citep{zhang2026deep} performs autonomous search over multi-granular video databases, and LVAgent~\citep{chen2025lvagent} introduces multi-round collaboration among MLLM agents. More recent approaches further combine active video inspection with reinforcement learning: VideoZoomer~\citep{ding2026videozoomer} learns coarse-to-fine temporal focusing, while VITAL~\citep{zhang2026thinking}, LongVT~\citep{yang2026longvt}, and Ego-R1~\citep{tian2026ego} train agents to dynamically invoke visual or retrieval tools for long-horizon reasoning. 
In summary, these methods primarily improve \emph{where} and \emph{when} to inspect the source video, while our work focuses on the complementary question of \emph{what should be persistently remembered} for subsequent retrieval.

\textbf{Memory-Based Long Video Understanding.}
Memory-based approaches retain video evidence for reuse across questions or as question-specific working memory. MovieChat and MA-LMM introduce sparse and online memory mechanisms~\citep{song2024moviechat,he2024ma}. EgoRAG organizes captions into hierarchical textual memory~\citep{yang2025egolife}, while HippoMM consolidates audiovisual events and combines summary retrieval with detailed recall~\citep{lin2026hippomm}. WorldMM~\citep{yeo2026worldmm} and MERIT~\citep{choi2026keep} construct and retrieve explicit memories using pretrained models without method-specific parameter updates. Training-based approaches include M3-Agent's separately trained memorization and control models~\citep{long2026seeing}, MemVid's memory-grounded clue generation optimized through SFT and DPO~\citep{yuan2025memory}, and Ego-R1's learned tool controller over hierarchical memory~\citep{tian2026ego}. VideoEvolve focuses on alternating the optimization of memory construction and retrieval policies, while keeping the constructed memory reusable across questions.

\textbf{Agentic Reinforcement Learning and Self-Evolving Agents.}
Reinforcement learning has become an effective mechanism for improving multimodal reasoning and decision-making, spanning visual perception~\citep{liu2025visual}, general multimodal reasoning~\citep{wang2026vl}, and video reasoning~\citep{feng2026video}. Agentic RL further couples policy optimization with multi-turn interaction and tool use, enabling models to learn adaptive behaviors in external environments~\citep{wang2025ragen,ding2026videozoomer,zhang2026thinking,yang2026longvt,tian2026ego}. Beyond single-policy optimization, recent self-evolving agents seek continual improvement from accumulated interaction experience: EvolveR~\citep{wu2025evolver} distills reusable strategies from past trajectories, SkillRL~\citep{xia2026skillrl} recursively evolves a hierarchical skill library together with the agent policy, Agent0~\citep{xia2025agent0} constructs progressively challenging curricula through autonomous interaction, and Evolving-RL~\citep{fan2026evolving} jointly optimizes experience extraction and utilization. 
Existing self-evolving agents mainly improve reasoning strategies, skills, and experience use. For long-video agents, however, \emph{what to remember} and \emph{how to retrieve} form an equally fundamental axis of evolution: what is preserved determines what can be retrieved, while retrieval utility in turn determines what is worth preserving.

\section{Method}

\subsection{Overview}

\begin{figure*}[t]
    \centering
    \includegraphics[width=\textwidth]{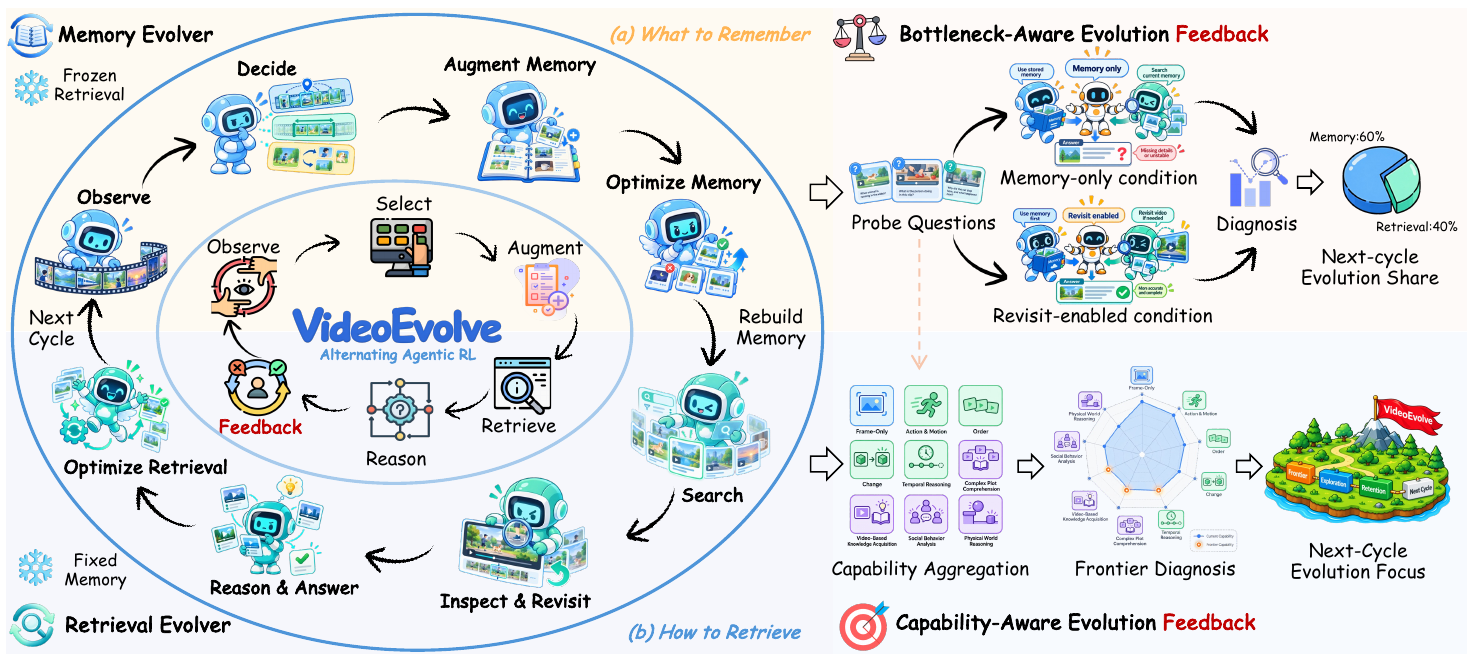}
    \vspace{-15pt}
    \caption{\textbf{Overview of the VideoEvolve.}
    VideoEvolve alternates between two stages: \textit{\textbf{(a) Memory Evolver}} learns what to remember with the Retrieval Evolver frozen, while \textit{\textbf{(b) Retrieval Evolver}} learns how to retrieve over the updated memory.
    \textit{\textbf{Bottleneck-Aware Evolution Feedback (BEF)}} and \textit{\textbf{Capability-Aware Evolution Feedback (CEF)} }further diagnose the current bottleneck and capability frontier to guide the next-cycle evolution. 
    }
    \vspace{-10pt}
    \label{fig:pipeline}
\end{figure*}

VideoEvolve (Figure~\ref{fig:pipeline}) comprises four tightly connected components:
(1) the \textbf{Memory Evolver} (Sec.~\ref{Sec.3.2}) learns \emph{what to remember} through selective memory augmentation;
(2) the \textbf{Retrieval Evolver} (Sec.~\ref{Sec.3.3}) learns \emph{how to retrieve} from the evolving memory;
(3) \textbf{Alternating Agentic RL} (Sec.~\ref{Sec.3.4}) co-evolves memory construction and retrieval through alternating optimization;
and (4) \textbf{Bottleneck-Aware Evolution Feedback (BEF)} and \textbf{Capability-Aware Evolution Feedback (CEF)} (Sec.~\ref{Sec.3.5}) diagnose the current bottleneck and capability frontier to guide the next cycle.

\subsection{Memory Evolver: Learning What to Remember}
\label{Sec.3.2}
The \textbf{Memory Evolver} (Figure~\ref{fig:pipeline}(a)) learns what information should be preserved through an \emph{Observe--Decide--Augment} process. We first construct a fixed hierarchical Base Memory $B$ from a coarse low-frame-rate overview, organized into \textit{Root--Super--Macro} nodes to provide broad temporal coverage and multi-granular retrieval anchors. The Base Memory remains fixed throughout evolution, while its complementary information acquisition is learned. Memory construction is question-agnostic, such that the Memory Evolver does not observe downstream questions or answers. The hierarchy, record schema, and retrieval organization are detailed in Appendix~\ref{app:base_memory}.

\textbf{Observe and Decide.}
At each construction step, the Memory Evolver observes the current memory state, construction history, and remaining budget. It first selects a Macro to inspect or terminates the trajectory, and then decides the temporal region, observation amount, and information focus:
\begin{equation}
u_t \sim \pi_C(\cdot \mid h_t^g;\theta), \qquad
z_t=(r_t,n_t,f_t)\sim\pi_C(\cdot \mid h_t^l,u_t;\theta),
\end{equation}
where $h_t^g$ denotes the global memory context, construction history, and remaining budget, while $h_t^l$ denotes the local context of the selected Macro. The variables $r_t$, $n_t$, and $f_t$ specify the region to inspect, the amount of visual observation, and the information focus, respectively.

\textbf{Augment and Rebuild Memory.}
According to the selected decision, the corresponding frames and subtitles are processed by a frozen multimodal Writer to produce structured, source-linked memory records. These records constitute a complementary Delta $\Delta_v^k$, whose compact summaries are further propagated to the associated Macro and Super nodes. The evolved memory at cycle $k$ is thus
\begin{equation}
M_v^k=\mathrm{Merge}(B_v,\Delta_v^k).
\end{equation}
After each Memory Evolver update, the memory is rebuilt from the same fixed Base using the updated construction policy rather than by accumulating historical Deltas. The policy optimization corresponding to \emph{Optimize Memory} in Figure~\ref{fig:pipeline} is formulated within the alternating Agentic RL framework in Sec.~\ref{Sec.3.4}.

\subsection{Retrieval Evolver: Learning How to Retrieve}
\label{Sec.3.3}
The \textbf{Retrieval Evolver} (Figure~\ref{fig:pipeline}(b)) learns to adaptively acquire and integrate information from the evolved memory for question answering. Given a question $q$ and the fixed memory $M$, it follows a \emph{Search--Inspect \& Revisit--Reason \& Answer} process to determine what information is needed and when retrieval should terminate.

\textbf{Search and Inspect \& Revisit.}
At each step, the Retrieval Evolver selects an action conditioned on the question, retrieved context, and interaction history:
\begin{equation}
a_t \sim \pi_R(\cdot \mid h_t^R;\phi),
\end{equation}
where $a_t$ can navigate the hierarchical memory, perform lexical or semantic search, access temporally localized content, inspect stored visual references, or selectively revisit the raw video. The policy therefore learns both where to search and whether additional visual observation is required.

\textbf{Reason and Answer.}
The retrieved information is progressively integrated until the Retrieval Evolver terminates the interaction and predicts the final answer $\hat{y}$. Before Agentic RL, the policy is initialized with supervised multi-turn tool-use trajectories to establish basic retrieval and interaction capabilities. The same policy can further operate under \emph{Memory-only} and \emph{Revisit-enabled} conditions for the evolution diagnosis introduced in Sec.~\ref{Sec.3.5}, and the corresponding retrieval policy optimization is formulated in Sec.~\ref{Sec.3.4}.

\textbf{Retrieval Tools.}
Inspired by MemDreamer~\citep{chen2026memdreamer}, our interface supports hierarchical navigation (\texttt{get\_macro\_events}, \texttt{get\_subgraph}), lexical--semantic and temporal search (\texttt{search\_nodes}, \texttt{search\_by\_time}), and inspection of paginated records and stored frames (\texttt{read\_document}, \texttt{get\_keyframes}) across Base and Delta memory.
When enabled, \texttt{view\_video} acquires new frames within budget.
Tool specifications and prompt templates are detailed in Appendices~\ref{app:tools} and~\ref{app:prompts}.

\subsection{Co-Evolving Memory and Retrieval via Agentic RL}
\label{Sec.3.4}
Memory construction constrains retrieval, whose outcomes reveal what is worth
preserving. We co-evolve both policies through \textit{Alternating Agentic
Reinforcement Learning (Agentic RL)}, optimizing one Evolver while keeping
the other side fixed.

\textbf{Alternating Co-Evolution.}
At cycle $k$, we first freeze the Retrieval Evolver $R^k$ and
\textit{\textbf{Optimize Memory}}. From the same Base Memory, the Memory Evolver
samples multiple construction trajectories, each producing a candidate memory.
The frozen $R^k$ evaluates these candidates on all questions for the video,
converting downstream answering outcomes into learning signals for memory
construction. After updating $C^k$ to $C^{k+1}$, we \textit{rebuild Memory}
from the same fixed Base using the updated policy. We then fix $M^{k+1}$ to
\textit{\textbf{Optimize Retrieval}}, yielding $R^{k+1}$:
\begin{equation}
(C^k,R^k,M^k)
\rightarrow
(C^{k+1},R^k,M^k)
\rightarrow
(C^{k+1},R^k,M^{k+1})
\rightarrow
(C^{k+1},R^{k+1},M^{k+1}).
\end{equation}
In this way, retrieval performance supervises what should be remembered, while
the rebuilt memory becomes the updated environment for subsequent retrieval
learning.

\textbf{Task-Specific Verifiable Rewards.}
The two Evolvers optimize different behaviors and therefore receive different
trajectory-level rewards. For a Memory Evolution trajectory $\tau_i^C$, let
$y_{ij}\in\{0,1\}$ indicate whether its candidate memory correctly supports
question $j$ under the frozen Retrieval Evolver, and let $y_j^{\mathrm{ref}}$
denote the corresponding result under the current reference memory. We define
\begin{equation}
R_C(\tau_i^C)
=
R_{\mathrm{rescue}}^{(i)}
-\lambda_{\mathrm{reg}}P_{\mathrm{regress}}^{(i)}
-\lambda_{\mathrm{inv}}P_{\mathrm{invalid}}^{(i)},
\end{equation}
where
\begin{equation}
\begin{aligned}
R_{\mathrm{rescue}}^{(i)}
&=
\frac{1}{N_q}\sum_{j=1}^{N_q}
\mathbf{1}[y_j^{\mathrm{ref}}=0 \land y_{ij}=1],\\
P_{\mathrm{regress}}^{(i)}
&=
\frac{1}{N_q}\sum_{j=1}^{N_q}
\mathbf{1}[y_j^{\mathrm{ref}}=1 \land y_{ij}=0].
\end{aligned}
\end{equation}
Here, $R_{\mathrm{rescue}}$ rewards newly rescued questions,
$P_{\mathrm{regress}}$ penalizes regressions on previously solved questions,
and $P_{\mathrm{invalid}}$ penalizes invalid construction trajectories. The
Memory Evolver is therefore encouraged to recover missing information without
sacrificing existing memory utility.

For a Retrieval Evolution trajectory $\tau_i^R$, the reward is directly tied to
question-answering success:
\begin{equation}
R_R(\tau_i^R)
=
R_{\mathrm{ans}}^{(i)}
-\lambda_{\mathrm{inv}}P_{\mathrm{invalid}}^{(i)},
\end{equation}
where $R_{\mathrm{ans}}^{(i)}=y_i\in\{0,1\}$ denotes final-answer correctness.
Thus, the Retrieval Evolver is directly reinforced for search and reasoning
trajectories that lead to correct answers.

\textbf{Quality-First Group-Relative Optimization.}
For either Evolver, multiple trajectories are sampled for the same training
instance and compared within a rollout group $\mathcal{G}$. Following
group-relative optimization, we first normalize the task reward:
\begin{equation}
A_{\mathrm{task}}^{(i)}
=
\frac{R_i-\mu_{\mathcal{G}}}
{\sigma_{\mathcal{G}}+\epsilon},
\end{equation}
where $\mu_{\mathcal{G}}$ and $\sigma_{\mathcal{G}}$ denote the mean and
standard deviation of rewards within the group.

Importantly, resource efficiency is not directly traded against task success.
Instead, it acts only as a tie-breaker among eligible trajectories with identical
task reward. Let $\mathcal{S}_i$ denote such trajectories and $E_i$ their
normalized resource cost. We define
\begin{equation}
A_{\mathrm{eff}}^{(i)}
=
\beta
\left(
\frac{1}{|\mathcal{S}_i|}
\sum_{j\in\mathcal{S}_i}E_j-E_i
\right),
\qquad
A^{(i)}
=
A_{\mathrm{task}}^{(i)}
+
A_{\mathrm{eff}}^{(i)},
\end{equation}
with $A_{\mathrm{eff}}^{(i)}=0$ when no eligible comparison exists. For Memory
Evolution, efficiency reflects additional visual observations and memory cost;
for Retrieval Evolution, it reflects tool use and additional raw-video
observation. Efficiency thus never overrides task correctness.

Finally, both Evolvers are optimized with the same clipped group-relative policy
objective:
\begin{equation}
\mathcal{L}_{\mathrm{RL}}
=
-\mathbb{E}_{i,t}
\left[
\min\left(
r_{i,t}A^{(i)},
\operatorname{clip}(r_{i,t},1-\epsilon_c,1+\epsilon_c)A^{(i)}
\right)
\right]
+
\beta_{\mathrm{KL}}D_{\mathrm{KL}},
\end{equation}
where $r_{i,t}$ denotes the policy ratio at generated token $t$,
$\epsilon_c$ controls the clipping range, and the KL term regularizes policy
updates. Resource costs, eligibility criteria, and numerical stabilization are detailed in
Appendix~\ref{app:objective}; parameter values are summarized in Table~\ref{tab:app_method_params}.

\subsection{Evolution Guidance via BEF and CEF}
\label{Sec.3.5}
Alternating Agentic RL defines how the Memory and Retrieval Evolvers are
optimized, but their relative training needs can shift across cycles.
Two complementary feedback mechanisms therefore determine \emph{how much}
each Evolver should evolve and \emph{where} next-cycle optimization should focus.

\textbf{Bottleneck-Aware Evolution Feedback (BEF).}
After each evolution cycle, we evaluate the current model on the same
\textbf{Probe Questions} under two paired conditions: \emph{Memory-only}, where
no new raw-video frames can be acquired, and \emph{Revisit-enabled}, where
selective raw-video observation is allowed. Let $m,v\in\{0,1\}$ denote answer
correctness under the two conditions, and let $f$ denote the number of newly
observed raw-video frames in the Revisit-enabled trajectory. We define
\begin{equation}
s_C=\mathbf{1}[m=0 \land v=1 \land f>0], \qquad
s_R=\mathbf{1}[v=0 \lor (m=0 \land v=1 \land f=0)].
\end{equation}
A Memory-only failure corrected through additional video observation contributes
to the memory-side demand, while unresolved failures or improvements without
new observation contribute to the retrieval-side demand. Aggregating these
paired signals yields $d_C$ and $d_R$, from which BEF computes the target Memory
Evolution share:
\begin{equation}
s_{\mathrm{target}}
=
\operatorname{clip}
\left(
\frac{d_C}{d_C+d_R},
s_{\min},s_{\max}
\right),
\qquad
s^{k+1}
=
\operatorname{clip}
\left(
s_{\mathrm{target}},
s^k-\delta_s,
s^k+\delta_s
\right).
\end{equation}
The complementary share is assigned to Retrieval Evolution. By dynamically reallocating optimization toward the currently more limiting side, BEF avoids a fixed training balance and allows the co-evolution process to continuously track shifting memory--retrieval bottlenecks across cycles.

\textbf{Capability-Aware Evolution Feedback (CEF).}
While BEF sets the two Evolvers' budgets, CEF prioritizes nine annotated
capabilities: Frame-Only, Action \& Motion, Order, Change, Temporal Reasoning,
Complex Plot Comprehension, Video-Based Knowledge Acquisition, Social Behavior
Analysis, and Physical World Reasoning (Appendix~\ref{app:cef_capabilities}).
For Evolver $p\in\{C,R\}$ and capability $c$, we aggregate paired diagnostic
signals into demand $d_{p,c}^{k}$ and update its smoothed state:
\begin{equation}
e_{p,c}^{k}
=
\alpha e_{p,c}^{k-1}
+
(1-\alpha)d_{p,c}^{k}.
\end{equation}
The next-cycle target capability distribution is
\begin{equation}
\widetilde w_{p,c}^{k+1}
=
\frac{\rho}{K}
+
(1-\rho)
\frac{e_{p,c}^{k}}
{\sum_{c'}e_{p,c'}^{k}},
\end{equation}
where $K$ counts capability categories and $\rho$ controls uniform coverage.
Before sampling, we project this target to obtain $w_{p,c}^{k+1}$ under the
coverage and rate bounds in Appendix~\ref{app:feedback}.

Per capability, \textbf{Frontier Diagnosis} mixes frontier (informative rollout variation),
exploration (underexplored cases), and retention (competence preservation) samples.
This \textbf{Next-Cycle Evolution Focus} samples videos/questions for $C$/$R$.
Fresh on-policy rollouts train $C$ to select observations and augment memory, and $R$
to acquire evidence and answer via alternating RL (Sec.~\ref{Sec.3.4}). Past trajectories
only guide sampling; capability labels are not policy inputs
(Appendix~\ref{app:cef_curriculum}; Table~\ref{tab:app_method_params}).

\subsection{Training and Inference}

\textbf{Training.}
Two-stage SFT initializes the Retrieval Evolver $R^0$ with context-conditioned
next-action supervision: Stage I teaches multi-turn tool use through high-quality
long-video trajectories; Stage II continues on a larger, more diverse
visual-trajectory set to broaden evidence seeking. The Memory Evolver skips
cold-start SFT. Each Agentic RL cycle updates the Memory Evolver with retrieval
frozen, rebuilds memory using its updated policy, then trains retrieval on fixed
memory. BEF and CEF set next-cycle evolution shares and capability focus; Base
Memory construction, the Writer, and retrieval tools remain fixed.

\textbf{Inference.}
The Memory Evolver builds evolved memory from each unseen video's fixed Base
for reuse across questions. The Retrieval Evolver answers independently with
optional raw-video revisiting. BEF and CEF are training-only; inference involves
no optimization.

\section{Experiment}
\subsection{Experimental Setting}

\textbf{Datasets and Evaluation Metrics.}
For the Retrieval Evolver's two-stage cold-start SFT, Stages I and II use
4,317 LongVT-derived~\citep{yang2026longvt} tool-use trajectories and 102,732
trajectories on LLaVA videos~\citep{zhang2025llavavideo}, respectively.
Both sets are generated by Qwen3.8-27B~\citep{qwen2026qwen38}.
After excluding audio-dependent samples, we use 514 videos
(2,056 questions) from Video-MME-v2~\citep{fu2026video} for Agentic RL.
A disjoint set of 128 videos (512 questions) supports BEF/CEF diagnosis
and scheduling without gradient updates.
We report multiple-choice accuracy on Video-MME~\citep{fu2025video},
LongVideoBench~\citep{wu2024longvideobench}, LVBench~\citep{wang2025lvbench},
MLVU~\citep{zhou2025mlvu}, and MMVU~\citep{zhao2025mmvu}.
Baseline categorization and source-specific evaluation settings are detailed in
Appendix~\ref{app:memory_protocols}.

\textbf{Implementation Details.}
For the training-based variant, the Memory Evolver, Retrieval Evolver, and frozen Writer use
Qwen3.5-4B~\citep{qwen2026qwen35}, Qwen3-VL-8B~\citep{bai2025qwen3}, and
Qwen3.8-27B~\citep{qwen2026qwen38}, respectively.
The training-free variant uses Qwen3.8-27B for both memory writing and retrieval.
Base Memory is constructed at 0.5 fps, and alternating Agentic RL uses
\texttt{verl}~\citep{sheng2025hybridflow}.
Training hyperparameters, memory construction settings, and tool budgets
are detailed in Appendices~\ref{app:training}, \ref{app:memory}, and~\ref{app:tools}, respectively.

\subsection{Comparisons with State-of-the-Art Methods}

\textbf{Comparison with Video-MLLMs.}
VideoEvolve establishes a clear advantage over existing open-source Video-MLLMs,
ranking first among the listed open-source methods on all six metrics across five
benchmarks (Table~\ref{tab:main_exp}). The gains are especially pronounced on
LongVideoBench and LVBench: VideoEvolve achieves \textbf{70.2\%} and \textbf{58.9\%},
surpassing the strongest listed open-source baselines, ParaVT and VideoZoomer,
by \textbf{9.8} and \textbf{17.4} percentage points, respectively. Notably, these
large margins are achieved over strong agentic baselines. VideoEvolve further
outperforms our tool-enabled Qwen3-VL-8B baseline by
14.7--28.6 points, marking a substantial advance over the off-the-shelf tool-enabled
configuration. Together with the single-Evolver ablations in Table~\ref{tab:ablation},
these results highlight the strength of our core design: co-evolving what is
remembered and how it is retrieved, rather than optimizing either side in isolation.

\begin{table*}[t]
\centering
\caption{\textbf{Video understanding comparison with Video-MLLMs.}
See Appendix~\ref{app:memory_protocols} for baseline sources and protocols; proprietary models retain their best settings.
Table~\ref{tab:memory_exp} compares memory-based methods. VideoEvolve-8B denotes an 8B Retrieval backbone.
\textbf{Bold}/\underline{underlined}: best/second-best listed open-source results; ``--'': unavailable.}
\vspace{-5pt}
\label{tab:main_exp}
\begingroup
\setlength{\tabcolsep}{5pt}
\renewcommand{\arraystretch}{1.08}
\resizebox{\textwidth}{!}{%
\begin{tabular}{l|cccccc}
\toprule[1pt]
\textbf{Model} &
\makecell{\textbf{VideoMME}\\w/o sub} &
\makecell{\textbf{VideoMME}\\w/ sub} &
\makecell{\textbf{LongVideo-}\\\textbf{Bench}} &
\textbf{LVBench} & \textbf{MLVU} & \textbf{MMVU} \\
\midrule\midrule
\multicolumn{7}{c}{\textbf{\textit{Proprietary Video-MLLMs: best-setting numbers from official reports}}} \\
\midrule
\textcolor{gray}{GPT-4o~\citep{hurst2024gpt}} &
\textcolor{gray}{71.9} & \textcolor{gray}{77.2} & \textcolor{gray}{66.7} &
\textcolor{gray}{34.7} & \textcolor{gray}{64.6} & \textcolor{gray}{66.7} \\
\textcolor{gray}{Gemini 1.5 Pro~\citep{team2024gemini}} &
\textcolor{gray}{75.0} & \textcolor{gray}{81.3} & \textcolor{gray}{64.4} &
\textcolor{gray}{33.1} & \textcolor{gray}{74.3} & \textcolor{gray}{65.8} \\
\midrule
\multicolumn{7}{c}{\textbf{\textit{Open Reasoning Video-MLLMs: reasoning-enhanced setting}}
(\texttt{<think>}$\rightarrow$\texttt{<answer>})} \\
\midrule
Video-R1-7B~\citep{feng2026video} & 57.6 & 66.0 & 57.4 & 36.9 & 61.6 & 61.3 \\
VideoChat-R1-7B~\citep{li2025videochatr1} & 50.4 & 58.2 & 49.2 & 23.8 & 58.7 & 65.0 \\
VideoRFT-7B~\citep{wang2025videorft} & 58.5 & 65.6 & 55.1 & 38.0 & 44.9 & 42.7 \\
Time-R1-7B~\citep{wang2025timer1} & 58.9 & 66.2 & 56.0 & 38.2 & 60.5 & 63.4 \\
ReWatch-R1-7B~\citep{zhang2026rewatchr1} & 58.8 & 65.0 & 53.6 & 38.5 & 60.1 & 59.8 \\
Video-Thinker-7B~\citep{wang2025videothinker} & 61.9 & 65.3 & 56.0 & -- & 65.2 & 64.5 \\
\midrule
\multicolumn{7}{c}{\textbf{\textit{Open Agentic Video-MLLMs: video-revisiting / tool-based inference}}} \\
\midrule
Qwen3-VL-8B (Baseline)~\citep{bai2025qwen3} & 49.5 & 52.7 & 45.1 & 30.3 & 55.1 & 60.4 \\
Conan-7B~\citep{ouyang2026conan} & 55.5 & 62.8 & 54.5 & 38.2 & 59.2 & 64.0 \\
LongVT-RFT-7B~\citep{yang2026longvt} & 59.5 & 66.0 & 54.7 & 37.9 & 59.4 & 63.4 \\
SAGE-7B~\citep{jain2026sage} & 44.1 & 52.4 & 37.4 & 31.8 & 49.7 & 55.7 \\
VideoZoomer-7B~\citep{ding2026videozoomer} & \underline{65.2} & -- & 57.7 & \underline{41.5} & \underline{68.8} & 61.6 \\
ParaVT-8B~\citep{yang2026paravt} & 62.1 & \underline{69.4} & \underline{60.4} & 39.8 & 65.0 & \underline{68.6} \\
\midrule
\rowcolor{blue!6}
\textbf{VideoEvolve-8B (Ours)} & \textbf{68.4} & \textbf{73.9} & \textbf{70.2} & \textbf{58.9} & \textbf{72.4} & \textbf{75.1} \\
\bottomrule[1pt]
\end{tabular}%
}
\endgroup
\vspace{-10pt}
\end{table*}

\textbf{Comparison with Memory-Based Methods.}
VideoEvolve also leads both groups of dedicated memory-based systems in
Table~\ref{tab:memory_exp}. With an 8B Retrieval Evolver, the trained
variant achieves \textbf{58.9\%} on LVBench and \textbf{67.1\%} on Video-MME (Long),
outperforming the listed training-based baselines. Its \textbf{14.5}-point margin
over MemVid, the strongest LVBench baseline in this group, is particularly
substantial, extending the advantage of our co-evolving system to methods that
already learn memory or retrieval components. The training-free Qwen3.8-27B
configuration likewise leads its group, reaching \textbf{76.7\%} and \textbf{78.8\%}
and surpassing MERIT-GPT by 4.9 and 1.1 percentage points, respectively.
This complementary result highlights the strength of our memory--retrieval design
even without task-specific parameter updates. Source-specific protocols and
configuration differences are detailed in Appendix~\ref{app:memory_protocols}.

\begin{table*}[t]
\centering
\caption{\textbf{Comparison with memory-based methods.}
Accuracy (\%) follows source-specific settings, detailed in Appendix~\ref{app:memory_protocols}. $^{\dagger}$ The memory model is fine-tuned on episodic annotations synthesized by Gemini 1.5 Pro and GPT-4o. ``--'' denotes an unreported result or unspecified model.}
\vspace{-5pt}
\label{tab:memory_exp}
\begingroup
\setlength{\tabcolsep}{4pt}
\renewcommand{\arraystretch}{1.12}
\resizebox{\textwidth}{!}{%
\begin{tabular}{lllcc}
\toprule
\textbf{Method} & \makecell[l]{\textbf{Memory}\\\textbf{Construction}} &
\textbf{Retrieval} & \textbf{LVBench} & \makecell{\textbf{Video-MME}\\\textbf{(Long)}} \\
\midrule
\multicolumn{5}{c}{\textbf{\textit{Training-free memory-based methods}}} \\
\midrule
EgoRAG~\citep{yang2025egolife} & -- & -- & 32.2 & 41.1 \\
HippoMM~\citep{lin2026hippomm} & Qwen2.5-VL & GPT-4o & 38.2 & 41.6 \\
WorldMM-GPT~\citep{yeo2026worldmm} & GPT-5-mini & GPT-5 & 61.9 & 76.6 \\
MERIT-GPT~\citep{choi2026keep} & GPT-5-mini & GPT-5 & \underline{71.8} & \underline{77.7} \\
\midrule
\rowcolor{blue!6}
\textbf{VideoEvolve (Ours)} & Qwen3.8-27B & Qwen3.8-27B & \textbf{76.7} & \textbf{78.8} \\
\midrule
\multicolumn{5}{c}{\textbf{\textit{Training-based memory-based methods}}} \\
\midrule
MemVid~\citep{yuan2025memory} & -- & Qwen2-VL-7B & \underline{44.4} & 57.1 \\
Ego-R1~\citep{tian2026ego} & Gemini 1.5 Pro & Qwen2.5-3B & 34.1 & \underline{64.9} \\
M3-Agent$^{\dagger}$~\citep{long2026seeing} & Qwen2.5-Omni-7B & Qwen3-32B & -- & 61.8 \\
\midrule
\rowcolor{blue!6}
\textbf{VideoEvolve (Ours)} & Qwen3.8-27B & Qwen3-VL-8B & \textbf{58.9} & \textbf{67.1} \\
\bottomrule
\end{tabular}%
}
\endgroup
\vspace{-15pt}
\end{table*}

\subsection{Ablation Studies}

Table~\ref{tab:ablation} examines three key designs of VideoEvolve; the full model
performs best across all seven metrics.
Cold-start and other design ablations appear in Appendix~\ref{sec:extended_ablation}.

\begin{table}[!t]
\centering
\caption{\textbf{Ablation of co-evolution and key components.}
Overall Video-MME uses the with-subtitles setting.}
\label{tab:ablation}
\vspace{-5pt}
\begingroup
\small
\begin{tblr}{
  width=\linewidth,
  colspec={X[l] Q[c,wd=1.15cm] Q[c,wd=1.0cm] Q[c,wd=1.27cm] Q[c,wd=1.2cm] Q[c,wd=1.35cm] Q[c,wd=1.05cm] Q[c,wd=1.15cm]},
  colsep=2.5pt, rowsep=1.3pt, cells={valign=m},
  row{1}={font=\bfseries}, row{2}={font=\itshape},
  row{3}={bg=blue!6},
  row{4,9,14}={font=\bfseries\itshape},
  cell{3}{2-8}={font=\bfseries},
  hline{1,Z}={0.8pt}, hline{3,4,9,14}={0.4pt},
  hline{2}={2-3,4-5}{0.3pt}
}
\SetCell[r=2]{l} Model & \SetCell[c=2]{c} Video-MME & & \SetCell[c=2]{c} LongVideoBench & & \SetCell[r=2]{c} LVBench & \SetCell[r=2]{c} MLVU & \SetCell[r=2]{c} MMVU \\
& overall & long & overall & long & & & \\
VideoEvolve & 73.9 & 67.1 & 70.2 & 65.5 & 58.9 & 72.4 & 75.1 \\
\SetCell[c=8]{l} (a) Co-evolution scheme: which Evolvers are trained and how to train & & & & & & & \\
Base (no evolution) & 66.8 & 57.2 & 62.1 & 57.3 & 50.3 & 64.7 & 67.5 \\
Memory evolution only & 68.8 & 62.1 & 63.0 & 58.8 & 51.1 & 67.3 & 69.2 \\
Retrieval evolution only & 69.9 & 62.5 & 66.3 & 61.0 & 52.7 & 68.8 & 70.1 \\
w/o alternating updates & 70.4 & 63.7 & 68.1 & 61.5 & 55.4 & 71.1 & 71.9 \\
\SetCell[c=8]{l} (b) Evolution guidance: training allocation and sample selection & & & & & & & \\
w/o BEF and CEF & 71.7 & 64.3 & 67.3 & 63.1 & 56.6 & 69.4 & 68.4 \\
w/o CEF & 72.8 & 66.3 & 68.4 & 64.3 & 57.9 & 70.7 & 73.2 \\
w/o BEF & 72.4 & 66.9 & 67.9 & 63.7 & 57.3 & 71.6 & 70.3 \\
w/o frontier diagnosis & 73.1 & 67.0 & 69.2 & 64.4 & 57.3 & 71.0 & 72.9 \\
\SetCell[c=8]{l} (c) Inference-time evidence access: raw-video revisiting & & & & & & & \\
w/o video revisiting & 72.3 & 65.5 & 68.8 & 64.1 & 57.5 & 71.6 & 73.9 \\
\end{tblr}
\endgroup
\vspace{-5pt}
\end{table}

\begin{table}[!t]
\centering
\caption{\textbf{Co-evolution dynamics on LVBench.}
$\Delta$Acc. is Revisit-enabled minus Memory-only accuracy.
Next-cycle training shares sum to 100\%.}
\label{tab:evolution_dynamics}
\vspace{-5pt}
\begingroup
\small
\begin{tblr}{
  width=\linewidth,
  colspec={Q[c,wd=0.8cm] Q[c,wd=2.3cm] Q[c,wd=2.6cm] Q[c,wd=1.3cm] X[c] X[c]},
  colsep=3pt, rowsep=2pt, cells={valign=m},
  row{1}={font=\bfseries}, row{2}={font=\itshape},
  row{6}={bg=blue!6,font=\bfseries},
  hline{1,Z}={0.8pt}, hline{3}={0.4pt},
  hline{2}={2-3,5-6}{0.3pt}
}
\SetCell[r=2]{c} Cycle & \SetCell[c=2]{c} Accuracy (\%) & & \SetCell[r=2]{c} {$\Delta$Acc.\\(pp)} & \SetCell[c=2]{c} Next-cycle training shares (\%) & \\
& Memory-only & Revisit-enabled & & Memory & Retrieval \\
0 & 45.6 & 50.3 & 4.7 & 50 & 50 \\
1 & 48.5 & 52.6 & 4.1 & 35 & 65 \\
2 & 54.9 & 57.1 & 2.2 & 45 & 55 \\
3 & 57.5 & 58.9 & 1.4 & 60 & 40 \\
4 & 58.1 & 58.6 & 0.5 & 45 & 55 \\
\end{tblr}
\endgroup
\vspace{-10pt}
\end{table}

\begin{samepage}
\textbf{Co-evolution.}
Joint evolution provides gains beyond single-sided optimization
(Table~\ref{tab:ablation}(a,b)). With BEF and CEF disabled in all three variants,
co-evolution outperforms either single-Evolver variant on six of seven metrics,
reaching 56.6\% on LVBench versus 51.1\% for memory-only and 52.7\% for
retrieval-only evolution. MMVU is the exception, addressed by feedback guidance
below. Removing alternating updates from the full system also lowers
LongVideoBench (Long) accuracy by 4.0 points. These comparisons support learning
both policies through alternating updates, rather than improving either side
in isolation.
\par
\end{samepage}

\begin{samepage}
\textbf{Evolution guidance.}
BEF and CEF close this gap and further strengthen co-evolution
(Table~\ref{tab:ablation}(b)). Their combination raises MMVU accuracy from 68.4\%
to 75.1\%, exceeding both BEF alone (73.2\%) and CEF alone (70.3\%). This
highlights the value of coordinating bottleneck-aware allocation with
capability-aware sampling. Disabling frontier diagnosis also reduces MMVU
accuracy by 2.2 points, supporting the selection of informative training instances
within each capability.
\par
\end{samepage}

\textbf{Video revisiting.}
Without raw-video revisiting, VideoEvolve still achieves 65.5\% on Video-MME (Long)
and 57.5\% on LVBench (Table~\ref{tab:ablation}(c)). Revisiting adds 0.8--1.6 points
on all seven metrics, complementing strong memory-based reasoning with selective
access to additional visual evidence.

\subsection{Analysis of Co-Evolution Dynamics}

Co-evolution delivers substantial gains even without raw-video access
(Table~\ref{tab:evolution_dynamics}). By Cycle 3, Memory-only accuracy rises
from 45.6\% to 57.5\%, surpassing the initial Revisit-enabled result by 7.2 points.
Revisit-enabled accuracy peaks at 58.9\%, while the gap narrows from 4.7 to 1.4
points. Cycle 4 further improves Memory-only accuracy but lowers Revisit-enabled
accuracy to 58.6\%. Together, these results validate the central premise of
VideoEvolve: effective long video reasoning benefits from jointly learning what
to preserve and how to use it.

\section{Conclusion}

In this work, we present VideoEvolve, a self-evolving framework that closes the
learning loop between memory construction and retrieval for long video understanding.
Through alternating Agentic RL, reasoning feedback guides what memory preserves,
while the updated memory shapes how retrieval learns to access evidence.
BEF and CEF steer this loop toward current bottlenecks and underdeveloped yet
learnable capabilities. Across five benchmarks, VideoEvolve outperforms the
evaluated open-source Video-MLLM baselines, and its trained and training-free
variants lead their respective memory-based comparisons. Ablations and cycle-wise
analysis further support this feedback-guided co-evolution. More broadly,
VideoEvolve reframes memory from a fixed record of video content into a learnable
interface between perception and reasoning, with its construction and use refined
together through task experience.

\bibliography{iclr2027_conference}
\bibliographystyle{iclr2027_conference}

\appendix
\clearpage
\begingroup
\raggedbottom
\setlength{\parskip}{4pt plus 1pt minus 1pt}
\setlength{\floatsep}{10pt plus 2pt minus 2pt}
\setlength{\textfloatsep}{10pt plus 2pt minus 2pt}
\setlength{\intextsep}{8pt plus 2pt minus 2pt}
\setlist[itemize,enumerate]{topsep=3pt,itemsep=2pt,parsep=0pt,partopsep=0pt}
\tcbset{before skip=5pt,after skip=5pt,left=2mm,right=2mm,top=1mm,bottom=1mm}
\SetTblrOuter[longtblr]{presep=8pt,postsep=8pt,headsep=4pt,footsep=3pt}
\SetTblrStyle{contfoot}{font=\small\itshape}
\SetTblrTemplate{caption}{plain}
\DeclareTblrTemplate{capcont}{appendix}{%
  \noindent\UseTblrTemplate{caption-tag}{default}\space
  \UseTblrTemplate{conthead-text}{default}\par}
\SetTblrTemplate{capcont}{appendix}

\section{APPENDIX}
\label{sec:impl_details}

This appendix provides implementation details, supplementary experimental results, and a discussion of limitations for VideoEvolve. Sections~\ref{app:memory} and~\ref{app:tools} describe memory construction and versioning, and the retrieval tools and interaction protocol, respectively. Section~\ref{app:data} presents data processing and two-stage cold-start training. Sections~\ref{app:objective} and~\ref{app:feedback} detail alternating updates and quality-first optimization, followed by BEF scheduling and CEF's capability definitions, diagnosis, and targeted training procedure. Section~\ref{app:training} summarizes training hyperparameters and runtime budgets, while Section~\ref{app:prompts} provides condensed prompt templates. Section~\ref{app:memory_protocols} clarifies evaluation settings and memory-based comparison protocols, and Section~\ref{sec:extended_ablation} reports extended ablation results. Finally, Section~\ref{app:limitations} discusses limitations and future work.

\subsection{Memory Construction and Versioning}
\label{app:memory}
\subsubsection{Fixed Hierarchical Base Memory}
\label{app:base_memory}
\paragraph{Fixed Base Memory.}
We uniformly sample the video at 0.5 fps and divide it into Macro intervals of at most 30 seconds, with the final interval clipped to the video duration. The frozen Writer processes the sampled visual evidence and available subtitles to produce local events, entities, OCR records, and summaries. A Base extraction request accepts at most 24 frames, resized to a maximum long edge of 896 pixels. Six-second CaptionUnits provide temporal anchors for subtitles and saved keyframes; they do not imply an additional visual-captioning call every six seconds. Macro-level extraction is consolidated before higher-level Super summaries are constructed, and the Root provides navigation over the hierarchy.

The Base is constructed once per video and is not rewritten during co-evolution. A policy version produces its own Delta from this same Base, and the published memory is the union of the Base and that version's accepted records. Historical Deltas are not recursively accumulated. All questions associated with a video share this memory, while question text, answer options, ground-truth answers, and capability labels are excluded from the construction policy's input. Entity records support local retrieval but do not constitute a guarantee of consistent identity tracking across the entire video.

\paragraph{Three-level organization.}
The Base separates a coarse-to-fine navigation hierarchy from its local evidence records and temporal index. Its three navigation levels are \textit{Root $\rightarrow$ Super $\rightarrow$ Macro}: one video-wide Root, a set of higher-level Super groups, and the Macro intervals that cover the video. Their roles and stored content are as follows.
\begin{enumerate}
    \item \textbf{Root: video-level overview.} The Root stores the full-video time range, duration, counts of Super and Macro nodes, a short description, and frequently mentioned entity names. Its description is formed deterministically by concatenating Super descriptions and retaining the first 500 characters; up to 20 entity names are selected by their frequency in Macro-level name lists. It provides an entry point into the video, not an exhaustive summary, and requires no additional Writer call.
    \item \textbf{Super: higher-level grouping.} Each Super stores a label, a narrative description, key entity names, member Macro IDs, and aggregated temporal anchors. After Macro extraction, the frozen Writer is prompted to group temporally adjacent and semantically related Macros. Invalid or repeated Macro assignments are removed, and omitted Macros are assigned through deterministic fallback grouping. A Super spans the earliest start and latest end of its members; unlike a Macro, it has no fixed duration.
    \item \textbf{Macro: local evidence container.} For a video of duration $T$, Macro $i$ covers $[30i,\min(30(i+1),T)]$ seconds, for $i=0,\ldots,\lceil T/30\rceil-1$. Each Macro stores its ID, interval, label, summary, detailed narrative, entity descriptors, event and OCR summaries, state changes, and references to CaptionUnits and saved keyframes. The detailed local records are retained separately and remain accessible through their Macro membership. Thus, upper-level summaries do not replace the underlying records.
\end{enumerate}

\paragraph{Local record schema.}
Within a Macro, \texttt{Event}, \texttt{Entity}, and \texttt{OCRText} are typed evidence records, not additional navigation levels. An Event stores its ID, description, event type, absolute time interval, subject, object, action, and references to local participant Entity IDs. An Entity stores a local ID, name, type, attributes, and description; visual-grounding metadata is retained when available. An OCRText record stores its ID, recognized text, type, description, and time interval. State changes and timestamped narrative details are also preserved in Macro fields. Event and OCR records carry time-basis metadata; an Entity without a localized visual timestamp is indexed over its containing Macro interval, not treated as a continuously tracked identity.

\paragraph{Hierarchy links and relation semantics.}
Explicit \texttt{SUBEVENT\_OF} links connect each Macro to its Super and each Super to the Root. Deterministic \texttt{BEFORE} chains connect successive Macro members within a Super and successive Super entries. The aggregation stage can additionally retain model-generated Super relations, such as causal or progression links; these are model inferences rather than independently verified facts. In contrast, the local Macro subgraph does not generate a separate semantic edge list: its compatibility \texttt{edges} field is empty. Participant references and relational descriptions reside in Event fields and text. The term ``subgraph'' therefore denotes the Macro's structured local records, not a fully populated entity--relation knowledge graph.

\paragraph{Subtitle and keyframe anchors.}
CaptionUnits form a separate six-second temporal grid, retaining IDs, start and end times, and keyframe IDs. Although the schema supports subtitle text, raw subtitles are not exposed to Retrieval in the evaluated configurations: subtitle information is available only through Writer-generated memory. A keyframe catalog maps each saved frame ID to its decoded timestamp and image path; here ``keyframe'' denotes a stored uniform sample, not a frame selected by a salience detector. Macro anchors connect these temporal units and images to the hierarchy. In the retrieval view, each CaptionUnit is assigned to one Macro by its temporal midpoint, avoiding duplicate ownership at interval boundaries. These units serve as temporal and image anchors, not raw-subtitle search documents. The associations complement the hierarchy rather than constituting another strict parent--child level.

\paragraph{Retrieval-facing representation.}
After bottom-up construction, a deterministic preprocessing step materializes compact Macro overviews, fine-grained documents, Macro-to-document membership lists, and the keyframe catalog. A local document uses the namespaced ID \texttt{macro\_id::local\_id}; CaptionUnits use \texttt{caption::caption\_id}. Documents retain their type, Macro and Super membership, time range, source reference, and original payload. Root and Super summaries provide initial navigation context and are not indexed as ordinary search documents; Macro overviews and nonempty local evidence provide coarse and fine textual search entries. Search text is separated from the stored record payload, so a concise hit does not discard the underlying details.

This representation supports navigation from a Super to its Macros with \texttt{get\_macro\_events}, inspection of a Macro's local records with \texttt{get\_subgraph}, and direct lexical--semantic or temporal entry through \texttt{search\_nodes} or \texttt{search\_by\_time}. The policy can then read paginated records or inspect associated stored images. It is not required to traverse every level from the Root for each question. Full tool arguments and access conditions are given in Appendix~\ref{app:tools}.

\subsubsection{Learned Augmentation and Versioning}
\label{app:memory_augmentation}
Figure~\ref{fig:base_to_evolved_memory} illustrates how the fixed Base becomes an augmented, reusable memory. The Base supplies initial evidence and navigation; the learned Controller selects additional observations, and the frozen Writer converts them into source-linked records.

\begin{figure}[!ht]
    \centering
    \includegraphics[width=\linewidth,pagebox=cropbox,clip]{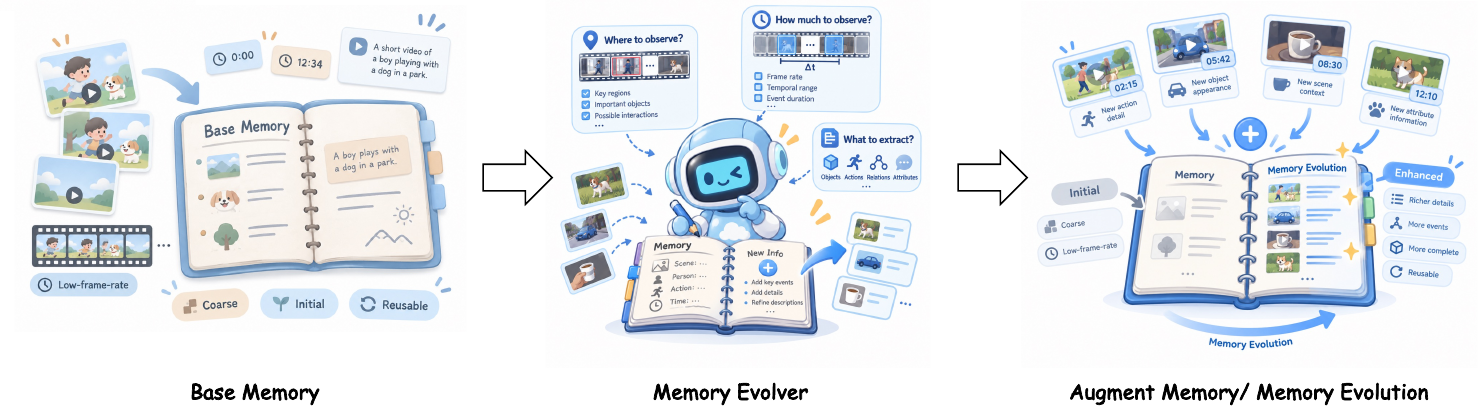}
    \caption{\textbf{From Base Memory to augmented memory.} Left: low-frame-rate sampling provides the fixed initial memory. Middle: the Memory Evolver selects where to observe, how many frames to request, and what information to extract. Right: accepted supplemental evidence enriches the retrieval view while preserving the Base. The timestamps and evidence cards are schematic, not a measured trajectory or a guarantee of complete coverage.}
    \label{fig:base_to_evolved_memory}
\end{figure}

\paragraph{Reading the three-stage process.}
Here, ``where'' denotes a Macro and temporal region, ``how much'' a discrete frame level within that region, and ``what'' an extraction focus. These choices direct observation toward evidence such as actions, object attributes, and interactions. Records passing the acceptance rules form the Delta: full records remain searchable, while bounded digests expose their content at Macro and Super levels. Thus, ``refine descriptions'' in the figure means supplementing the retrieval view, not overwriting the original Base. The resulting Base-plus-Delta memory and its matching index are shared by questions about the same video.

\paragraph{Augmentation versus policy evolution.}
Within one construction trajectory, the Controller accumulates evidence under fixed budgets without updating its parameters. Across training cycles, the frozen Retrieval policy evaluates candidate memories on downstream questions, and quality-first RL updates the Controller's observation decisions (Appendix~\ref{app:objective}). The updated Controller then reconstructs each training video's memory from the same Base, producing a new Delta and index rather than appending historical Deltas. Evolution therefore concerns the learned construction policy and the memories it produces; gradients do not pass through memory records or the frozen Writer. At inference, the learned policy constructs a question-agnostic memory without online policy optimization.

\paragraph{Two-stage construction actions.}
The first Controller decision selects an actionable Macro or \texttt{STOP}. Its context contains Base summaries, bounded Delta digests, observation history, and remaining budgets. The second decision receives details only for the selected Macro and chooses a region, a frame level, and an extraction focus. The regions are \texttt{full}, \texttt{head}, \texttt{middle}, \texttt{tail}, and \texttt{boundary}. The three interior regions divide the Macro into thirds; the boundary region spans 20 seconds around its end, clipped to the video, and is absent for the final Macro. Either stage may stop the entire construction trajectory; stopping at the second stage does not return to Macro selection.

Frame levels are $\{2,4,8,16\}$. The focus is one or two of \texttt{action\_\allowbreak state}, \texttt{appearance\_\allowbreak spatial}, and \texttt{text\_\allowbreak alignment}, or \texttt{general} alone, giving seven combinations. Each region permits at most two observations, and the exact tuple of Macro, region, frame level, and canonicalized focus cannot repeat. An observation consumes a visit even if the Writer returns no new record. Frame targets are nested across levels, mapped to actual source-frame timestamps within the requested window, and deduplicated within a request. Consequently, the number of decoded frames can be smaller than the selected level.

\paragraph{Separate observation and text budgets.}
For duration $T$ in seconds, we set the supplemental frame budget to
\begin{equation}
 b_F(T)=\begin{cases}
 32,&0<T\leq300,\\
 64,&300<T\leq900,\\
 96,&900<T\leq1800,\\
 128,&T>1800.
 \end{cases}
 \label{eq:app_frame_budget}
\end{equation}
The accepted supplemental description budget is $b_T=4{,}096$ tokens per video, with at most 256 tokens per step. Multiple focuses share one observation and text allowance. Frame accounting uses images actually delivered to the Writer, including additional deliveries for format repair. Transport retries are logged separately as service costs and do not create new policy actions. Base construction, supplemental observation, and question-time raw-video revisiting have separate budgets; none of these frame counts is a complete measure of end-to-end compute.

\paragraph{Source-linked acceptance.}
Each proposed record contains a description, an absolute time interval, referenced frame and subtitle IDs, and a relation flag indicating additional or conflicting information. The executor checks field types, finite and legal timestamps, source-ID existence, correspondence to the supplied observation, and text length. Records must cite supplied evidence. After these structural checks, descriptions are deduplicated by lowercasing and collapsing whitespace. Records explicitly marked as conflicts by the Writer are isolated rather than used to overwrite the Base. There is no additional learned verifier or semantic audit: valid provenance and formatting do not establish that every claim is true or that all semantic conflicts have been detected.

\paragraph{Making the Delta retrievable.}
Accepted records remain individually searchable as fine-grained Events. Deterministic, length-bounded digests expose these records at their associated Macro and Super nodes, with evidence IDs retained and the original Base summaries stored separately. Boundary evidence is assigned to a Macro using its temporal midpoint. Overlong records can be represented by temporal pointers, and the digest need not contain every accepted detail. The complete records remain available through retrieval. Each published Base-plus-Delta view is paired with its own index; a Base index must not be reused as though it indexed an Enhanced memory.

\subsection{Retrieval Tools and Interaction Protocol}
\label{app:tools}
\paragraph{Access conditions.}
Memory-only inference allows access to textual memory records and stored keyframes, but no newly acquired source-video frames. Revisit-enabled inference additionally exposes \texttt{view\_video}; enabling it does not force a revisit. We allow 64 raw-video frames per question in total and at most 16 per call, resized to a maximum long edge of 896 pixels. Stored-frame inspection does not consume this raw-video allowance, although it consumes a tool call and contributes multimodal input to the policy.

\begin{table}[htbp]
\centering
\caption{Retrieval tool interface. All temporal arguments are absolute video seconds; availability and parameter bounds are enforced by the runtime schema.}
\label{tab:app_tools}
\footnotesize
\begin{tblr}{width=\linewidth,colspec={Q[l,wd=3.05cm] X[l]},colsep=5pt,rowsep=4pt,row{1}={font=\bfseries},hline{1,Z}={0.8pt},hline{2}={0.4pt}}
Tool & Arguments and returned evidence \\
\texttt{get\_macro\_events} & No filter, or \texttt{super\_id}, or \texttt{macro\_ids}; enumerates Macro summaries and intervals. Filters are mutually exclusive. \\
\texttt{get\_subgraph} & \texttt{macro\_id}; returns local event, entity, and OCR records. \\
\texttt{search\_nodes} & \texttt{query}; optional \texttt{node\_types}, \texttt{top\_k}, and \texttt{weight\_preset}; retrieves candidate nodes. \\
\texttt{search\_by\_time} & \texttt{start\_sec}, \texttt{end\_sec}, with start $<$ end; locates temporally overlapping evidence. \\
\texttt{get\_keyframes} & \texttt{doc\_ids}, \texttt{max\_images}; optional paired start/end times; returns only stored images in the requested scope. \\
\texttt{read\_document} & \texttt{doc\_ids}, \texttt{offset}, \texttt{limit}; reads paginated document content. \\
\texttt{view\_video} & Start/end times and \texttt{frames}; acquires source-video frames only when revisiting is enabled. \\
\end{tblr}
\end{table}

\paragraph{Evidence retrieval and temporal grounding.}
Node search combines lexical BM25 and normalized dense-vector cosine similarity through weighted reciprocal-rank fusion. The embedding service uses Qwen3-Embedding-8B with 4,096-dimensional vectors and is held fixed during policy learning. Search hits and Root/Super summaries are navigation cues, not proof that every query condition is satisfied. The policy can inspect full records, page through truncated documents, and check stored images. Stored-frame requests with equal start and end times target that exact timestamp and may return no image; the executor does not silently substitute a nearby frame outside the requested interval.

\paragraph{Decision and stopping rules.}
Each interaction has 12 exploration slots, followed, if needed, by one final-only decision. During exploration, the policy may call a tool or answer; it may answer immediately when the initial evidence suffices. Abstention is available only in the final-only decision, which allows no further tool calls. Invalid intermediate actions execute no tool but consume a slot and return explicit validation feedback. The same policy performs evidence acquisition, reasoning, and answering; no separate answerer or verifier is invoked. Final citations must refer to observed \texttt{ev\_*} records, although citation validity alone does not establish semantic support.

\paragraph{Constrained decoding and context.}
Controller decisions use an enumeration of complete legal JSON actions, including \texttt{STOP}, so joint constraints on region, frames, and focus cannot be bypassed by combining individually valid fields. Retrieval uses the decision-specific JSON schema, available tools, and observed evidence IDs. RL rollouts retain the actual input and constraint state for each decision. Old-policy, updated-policy, and reference log probabilities are computed with the same legal-token masks and sampling temperature. Tool outputs, images, and earlier observations are context rather than newly generated Actor targets.

\subsection{Data Processing and Two-Stage Cold Start}
\label{app:data}
\paragraph{Stage I.}
The LongVT-derived cold-start set contains 4,317 accepted trajectories from 4,205 videos, expanded into 18,932 decision samples. This includes 56 validated retry trajectories added to the initial 4,261. Each sample retains the observation and interaction history available at that decision, while supervising only the final assistant response. The stage establishes the tool-use protocol; it does not supervise the Memory Evolver.

\paragraph{Stage II filtering.}
The larger teacher pool on LLaVA videos is filtered for correct final answers, at least one tool action, schema-conformant original JSON responses, valid final references, and available image files. Zero-tool answers, repaired trajectories, and cases flagged for further review are kept separate from this strict training pool. This selection criterion encourages tool-grounded demonstrations; it does not forbid zero-tool answers at inference or amount to a human semantic verification of every rationale.

Within the strict pool, trajectories are near-duplicates when they share a video and the complete tool path and their question-word sets have Jaccard similarity at least 0.8. Removing 190 such trajectories reduces the pool from 102,922 to 102,732 complete trajectories across 9,995 videos, yielding 349,717 decision samples. We use this full deduplicated set, not the earlier half-data selection. All decision turns in each retained trajectory are preserved; Stage II starts from the final Stage-I checkpoint rather than from the half-data Stage-II model.

\paragraph{Input integrity and separation.}
Both stages mask prompts and history from the supervised loss and disable sequence packing. Ground-truth labels, correctness flags, provenance metadata, and filtering reasons are not appended to Actor prompts. Stage-II full-input tokenization totals 2,072,137,576 tokens with a maximum sample length of 25,769; these totals include context and are not counts of supervised output tokens. All 349,717 samples pass the recorded no-truncation check, with an additional multimodal-processor cross-check on 128 image-containing samples. Video-ID overlap checks against the available evaluation lists find no overlap; this is ID-based decontamination, not a guarantee against all content-level duplicates. Neither SFT stage holds out an internal validation split.

\paragraph{RL and diagnostic pools.}
After removing audio-dependent questions from Video-MME-v2, we retain only videos with four complete questions: 514 videos and 2,056 questions for training, and 128 videos and 512 questions for diagnosis. Video and sample IDs are disjoint between the two pools. The diagnostic set is reused for BEF/CEF scheduling and is therefore a development probe, not an independent final test set. Capability labels and answers are available only to scoring and scheduling components. The training questions remain fixed; the framework does not generate new questions online.

\subsection{Alternating Updates and Quality-First Optimization}
\label{app:objective}
\paragraph{Construction-first update sequence.}
At cycle $k$, each construction group contains eight candidate trajectories for one video. Candidates start from the fixed Base, while their reward reference is the currently published Enhanced memory $M^k$, not the Base. The frozen $R^k$ scores each candidate and the reference on all four training questions in memory-only mode. After the Controller update, $C^{k+1}$ reconstructs a unified memory and its index for every training video. This is a fresh construction pass, not publication of the best of the eight sampled candidates. Retrieval training then samples four trajectories per question with raw-video access enabled, holding the rebuilt memory fixed. No gradients pass through the Writer, memory files, index, or frozen scorer.

\paragraph{Quality signals and resource costs.}
The rewards in Sec.~\ref{Sec.3.4} use $\lambda_{\mathrm{reg}}=1$ and $\lambda_{\mathrm{inv}}=0.2$. For valid construction trajectories, the rescue-minus-regression reward is then exactly candidate accuracy minus reference accuracy; an extra absolute-accuracy term is not added. Resource costs are computed separately:
\begin{align}
 E_C^{(i)}&=\frac{2\min(F_i^C/b_F(T),1)+\min(L_i/b_T,1)}{3},\\
 E_R^{(i)}&=\frac{\min(N_i/B_N,1)+\min(F_i^R/B_F^R,1)}{2}.
 \label{eq:app_costs}
\end{align}
Here $F_i^C$ counts supplemental Writer frame inputs, $L_i$ counts accepted description tokens, $N_i$ counts executed retrieval tools, and $F_i^R$ counts newly acquired raw-video frames. The normalizers are $b_T=4{,}096$, $B_N=12$, and $B_F^R=64$, with $b_F(T)$ defined in Eq.~\ref{eq:app_frame_budget}. In memory-only mode the raw-frame term is defined as zero, without dividing by a disabled budget. These bounded costs lie in $[0,1]$ and are proxies for selected resources, not measured latency or GPU hours.

\paragraph{Eligibility and tie-breaking.}
A Retrieval trajectory is efficiency-eligible only when its final answer is valid and correct. A construction trajectory must be valid, have utility no lower than the current reference, and have nonnegative task reward. For an eligible trajectory $i$, $\mathcal{S}_i$ contains the eligible members of its group with the same task reward up to the numerical tolerance defined below. The efficiency term is nonzero only when $|\mathcal{S}_i|\geq2$, and uses $\beta=0.05$. The final advantage is the task advantage plus the efficiency term, with no second normalization and no subtraction of cost before task normalization.

Thus, an all-incorrect or all-abstaining Retrieval group receives no positive preference merely for saving tools. Construction groups below the reference may still provide task-quality learning signals when their rewards differ, but receive no efficiency preference. Since $E_i\in[0,1]$, the tie-breaking contribution has magnitude at most 0.05. The common reference cancels from within-group quality differences when $\lambda_{\mathrm{reg}}=1$; its additional roles are to define rescue/regression outcomes and eligibility, not to introduce new candidate rankings by itself.

\paragraph{Numerical stabilization.}
The constant $\epsilon$ in Sec.~\ref{Sec.3.4} is $10^{-8}$. We compute $\sigma_{\mathcal{G}}$ as the population standard deviation over complete trajectories, not over individual decision turns. The same constant stabilizes the denominator and determines when the task-quality signal is numerically degenerate:
\begin{equation}
 A_{\mathrm{task}}^{(i)}=
 \begin{cases}
 \dfrac{R_i-\mu_{\mathcal{G}}}{\sigma_{\mathcal{G}}+\epsilon},&\sigma_{\mathcal{G}}>\epsilon,\\[3pt]
 0,&\sigma_{\mathcal{G}}\leq\epsilon,
 \end{cases}
 \qquad \epsilon=10^{-8}.
 \label{eq:app_adv_stability}
\end{equation}
The implementation also uses $|R_i-R_j|\leq\epsilon$ to identify equal-quality peers for efficiency tie-breaking. A zero task advantage does not by itself discard a group: eligible equal-quality trajectories can still receive an efficiency signal, and only an all-zero final advantage vector causes the update to be skipped. The numerical tolerance $\epsilon$ is distinct from the policy-clipping parameter $\epsilon_c$; likewise, the efficiency coefficient $\beta$ is distinct from the KL coefficient $\beta_{\mathrm{KL}}$.

\paragraph{Episode-balanced Actor loss.}
Let $\mathcal{Y}_i$ contain all Actor-generated output tokens across the decisions of trajectory $i$. The expectation in the main-text clipped objective is implemented by first averaging over $\mathcal{Y}_i$ and then over trajectories:
\begin{equation}
 \mathcal{L}_{\mathrm{policy}}=-\frac{1}{|\mathcal{G}|}\sum_{i\in\mathcal{G}}
 \frac{1}{|\mathcal{Y}_i|}\sum_{t\in\mathcal{Y}_i}
 \min\!\left(r_{i,t}A^{(i)},\operatorname{clip}(r_{i,t},1-\epsilon_c,1+\epsilon_c)A^{(i)}\right).
\end{equation}
The same delayed advantage supervises the construction trajectory's two-stage decisions or the Retrieval trajectory's successive decisions. Prompts, tool responses, image positions, and padding are excluded from the output loss mask. We use $\epsilon_c=0.2$ and add KL regularization with coefficient $\beta_{\mathrm{KL}}=0.02$ to the loss; the KL term is not also added to the reward. A group with all-zero final advantages is skipped. Missing trajectories or infrastructure failures invalidate the whole group rather than becoming negative policy rewards, whereas observed invalid policy terminations remain scoreable outcomes.

\subsection{BEF and CEF Scheduling Details}
\label{app:feedback}
\subsubsection{Paired Diagnosis and BEF Budgets}
\paragraph{Paired diagnostic panel.}
After both policy updates, the new Controller constructs probe memories without seeing the probe questions. The new Retrieval policy is evaluated on the same 512 questions under both access modes using seeds $\{17,29,43,71\}$ and temperature 0.7, yielding $512\times2\times4=4{,}096$ diagnostic trajectories. This is a panel of the current memory--retrieval pair, not a cross-product of old and new systems. Signals $s_C$ and $s_R$ from Sec.~\ref{Sec.3.5} are averaged over seeds within a question and then over questions; overall demands preserve question-count weighting across capabilities.

BEF uses the actual raw-frame count, not merely the availability of \texttt{view\_video}. A memory-only failure rescued with positive raw-frame use contributes to construction demand; an unresolved revisit-enabled failure or a rescue without new frames contributes to retrieval demand. These are operational allocation signals, not causal proofs of missing memory or defective reasoning. In particular, failure in both conditions does not establish that memory construction is already adequate.

\paragraph{From shares to executable groups.}
We initialize the construction share at $s^0=0.5$, set the target bounds to $s_{\min}=0.25$ and $s_{\max}=0.75$, and limit the per-cycle change with $\delta_s=0.15$. These are the parameters in Sec.~\ref{Sec.3.5}; the Retrieval share is $1-s$. If $d_C+d_R=0$, the previous share is retained. With $N_C=514$ video groups and $N_R=2{,}056$ question groups in one full pass, the next-cycle group counts are
\begin{equation}
 G_C^{k+1}=\operatorname{round}(2s^{k+1}N_C),\qquad
 G_R^{k+1}=\operatorname{round}(2(1-s^{k+1})N_R).
\end{equation}
At equal shares, each side receives one full pass: 4,112 construction candidates and 8,224 Retrieval trajectories. With one scorer seed, evaluating the reference and eight candidates on four questions requires 18,504 frozen-scorer trajectories for the construction pass. The shares therefore describe normalized training passes, not equal wall-clock time or GPU cost. Updating fewer construction groups does not eliminate the subsequent full memory/index reconstruction pass.

\subsubsection{Capability Taxonomy and Label Semantics}
\label{app:cef_capabilities}
CEF uses the nine non-audio parent categories of Video-MME-v2~\citep{fu2026video}, retained as the question-level \texttt{second\_head} labels in our RL and probe pools. The finer \texttt{third\_head} annotations are preserved but are not independently scheduled. We neither relabel questions nor assign a video's final-question category to all its questions. The categories specify the evidence and reasoning demands of questions, rather than separate model modules:
\begin{enumerate}
    \item \textbf{Frame-Only.} Recognizing entities, attributes, scenes, and visible text; counting objects; and performing simple calculations from visual information. Evidence may be aggregated across frames, but the answer does not require their temporal order. Thus, this label does not impose a single-frame input limit.
    \item \textbf{Action \& Motion.} Distinguishing fine-grained actions, counting action repetitions, locating actions in time, and analyzing trajectories and motion properties. These tasks depend on visual dynamics across time rather than static appearance alone.
    \item \textbf{Order.} Recovering the temporal order of entity appearances or event sequences. The distinguishing requirement is which appearance or event precedes another, rather than merely recognizing the participating entities or events.
    \item \textbf{Change.} Comparing entity existence, attributes, or scene states across time, and detecting periodic or cyclic patterns. These questions require relating observations at different moments to identify what changed or recurred.
    \item \textbf{Temporal Reasoning.} Inferring video-grounded cause--effect relations and predicting subsequent events from observed dynamics. Unlike Order, this category requires an inference about why an event occurs or what is likely to follow, not only chronological sorting.
    \item \textbf{Complex Plot Comprehension.} Integrating narrative evidence to identify turning points, infer unstated implications, interpret symbolism or metaphor, and understand higher-order storytelling structure. Relevant clues may be distributed across multiple scenes.
    \item \textbf{Video-Based Knowledge Acquisition.} Understanding domain knowledge or practical procedures presented in the video, including demonstrated steps, conditions, and error checks. Answers must depend on the presented content rather than generic background knowledge alone.
    \item \textbf{Social Behavior Analysis.} Inferring individual intentions, emotions, or mental states, and analyzing interactions between two people or group-level roles, cooperation, and conflict from contextual evidence across time.
    \item \textbf{Physical World Reasoning.} Reasoning about entity persistence, three-dimensional spatial relations, grounded counterfactuals, and counterintuitive physical phenomena. The task is to connect observations to physical continuity or constraints, rather than only report visible attributes.
\end{enumerate}

These labels remain outside both Actor prompts. In particular, the Memory Evolver receives neither a question nor a requested capability: it constructs one reusable memory per video. CEF changes which labeled training instances are sampled, without adding capability-specific heads, prompts, rewards, or new questions.

\subsubsection{From Capability Diagnosis to Targeted Policy Updates}
\label{app:cef_curriculum}
\paragraph{Capability-specific demand.}
Let $\mathcal{P}_c$ be the probe questions with label $c$ and $\mathcal{S}$ the paired seed set. For each Evolver $p\in\{C,R\}$, CEF aggregates the signals defined in Sec.~\ref{Sec.3.5} as
\begin{equation}
 d_{p,c}^{k}=\frac{1}{|\mathcal{P}_c|}\sum_{q\in\mathcal{P}_c}
 \frac{1}{|\mathcal{S}|}\sum_{z\in\mathcal{S}}s_p^{k}(q,z).
 \label{eq:app_cef_demand}
\end{equation}
Thus, raw-frame-assisted rescues within a category contribute to its construction demand, whereas unresolved revisit-enabled errors and rescues without new frames contribute to its retrieval demand. The two estimates and sampling distributions are maintained separately: one capability can require different training emphasis on the two sides. These are the same operational proxies used by BEF, now conditioned on capability, not independent measurements of an isolated latent skill.

\paragraph{Capability state and coverage.}
The initial sufficiently supported probe initializes each demand estimate; subsequent estimates use EMA coefficient $\alpha=0.5$. A missing observation or support from fewer than ten distinct probe videos retains the previous estimate, or zero if none is available. We set $\rho=0.3$: each Evolver's target distribution mixes 30\% uniform coverage over $K=9$ categories with 70\% normalized smoothed demand. If all demand estimates are zero, the target is uniform.

The distribution in Sec.~\ref{Sec.3.5} specifies the target $\widetilde w_{p,c}^{k+1}$. Before sampling, the implementation projects it onto the probability simplex with simultaneous coverage and rate bounds to obtain $w_{p,c}^{k+1}$:
\begin{align}
 \ell_{p,c}&=\max(\rho/K,\,0.5w_{p,c}^{k}),\qquad
 u_{p,c}=\max(\ell_{p,c},\,2w_{p,c}^{k}),\\
 w_{p,c}^{k+1}&=\operatorname{clip}(\widetilde w_{p,c}^{k+1}+\delta_p,\ell_{p,c},u_{p,c}),
 \qquad \sum_c w_{p,c}^{k+1}=1.
\end{align}
Here $K=9$, $\rho=0.3$, and the shared offset $\delta_p$ is computed by bisection rather than chosen as a hyperparameter. This projection preserves all bounds jointly; independently clipping each weight and then renormalizing need not do so. BEF controls the numbers of training groups, while CEF controls their within-role capability composition.

\paragraph{Role-specific training instances.}
Cycle 0 makes one full pass over all training videos for construction and all training questions for retrieval. Subsequent cycles use the feedback distributions within BEF's group budgets. For Retrieval, we first draw $c$ according to $w_{R,c}^{k+1}$ and then draw a training question with that label. For construction, we draw $c$ according to $w_{C,c}^{k+1}$ and select a video containing at least one question of that category. A video can therefore belong to several construction buckets. Each video is selected at most twice per cycle; once this cap is reached it is excluded, and capability sampling uses the remaining weights restricted to categories with eligible videos.

\paragraph{Frontier, exploration, and retention.}
Within the selected capability bucket, a branch is drawn with probabilities 0.6, 0.2, and 0.2, respectively; these are sampling probabilities, not exact finite-batch quotas. Retrieval frontier questions have at least two valid historical revisit-enabled trajectories with both correct and incorrect outcomes; retention questions have only correct valid outcomes. Construction frontier videos have a four-question utility range greater than $10^{-8}$ within a complete, valid candidate group evaluated by the same frozen scorer. Retention videos have at least one complete valid group in which every candidate answers all four questions correctly. Comparisons across different scorer versions are not used to establish construction frontier membership.

Exploration samples uniformly from the full eligible bucket, including previously unseen, failed, and successful instances. A frontier or retention branch with no candidates falls back to this full bucket; otherwise sampling is uniform within the chosen branch. Construction frontier takes precedence over retention when both apply. This mixture emphasizes observed variation while maintaining access to unsolved cases and opportunities to retain successful behavior. Historical frontier membership is an empirical indicator of useful variation, not a guarantee of learnability.

\paragraph{How the selected capabilities are trained.}
Every selected instance generates fresh on-policy rollouts; previous-cycle outcomes determine sampling only, and probe examples never enter the gradient-training pool. A selected construction video produces eight Controller trajectories from the fixed Base. The frozen Retrieval policy scores their memories and the current reference on all four associated questions with equal weight, even when the video was selected for only one capability. The group-relative, quality-first objective in Sec.~\ref{Sec.3.4} updates the Controller's choices of where to observe, how many frames to request, what to extract, and when to stop. Capability targeting thus changes exposure to videos with particular evidence demands, not the four-question reward or the frozen Writer.

After updating the Controller, we rebuild all training memories and indexes. Each selected Retrieval question generates four revisit-enabled trajectories over the fixed rebuilt memory, and quality-first RL updates evidence acquisition, reasoning, and answering. A new paired probe then sets the next-cycle focus. For Order, construction demand prioritizes videos containing ordering questions, whereas retrieval demand prioritizes those questions, without prescribing tool sequences. This is a mechanism illustration, not a measured capability gain. BEF and CEF remain training-only.

\subsection{Training Hyperparameters and Runtime Budgets}
\label{app:training}
Table~\ref{tab:app_method_params} maps the method's parameter symbols to their meanings, values used in our experiments, and main-text sections. The BEF shares and CEF sampling weights themselves evolve during training. The quantities $N_q$ and $K$ describe the question panel and capability taxonomy rather than tunable loss coefficients. Construction and retrieval budget semantics are detailed in Sections~\ref{app:memory} and~\ref{app:tools}, numerical stabilization in Section~\ref{app:objective}, and feedback scheduling in Section~\ref{app:feedback}.

\begin{longtblr}[
caption={Method parameters used in our experiments. C and R denote construction and retrieval. Supplemental frame budgets depend on video duration as specified in Eq.~\ref{eq:app_frame_budget}.},
label={tab:app_method_params}
]{width=\linewidth,colspec={Q[c,wd=1.4cm] X[l] Q[c,wd=2.4cm] Q[c,wd=1.15cm]},colsep=4pt,rowsep=1.5pt,cells={valign=m,font=\small},rowhead=1,row{1}={font=\small\bfseries},hline{1,Z}={0.8pt},hline{2,11,21}={0.4pt},hborder{3,12,22}={pagebreak=no}}
Symbol & Meaning & Value & Method \\
\SetCell[c=4]{l,font=\small\bfseries\itshape} Quality-first optimization & & & \\
$\lambda_{\mathrm{reg}}$ & Regression penalty & $1$ & \ref{Sec.3.4} \\
$\lambda_{\mathrm{inv}}$ & Invalid-terminal penalty & $0.2$ & \ref{Sec.3.4} \\
$\beta$ & Efficiency tie-breaking coefficient & $0.05$ & \ref{Sec.3.4} \\
$\epsilon$ & Numerical tolerance & $10^{-8}$ & \ref{Sec.3.4} \\
$\epsilon_c$ & Policy-ratio clipping range & $0.2$ & \ref{Sec.3.4} \\
$\beta_{\mathrm{KL}}$ & KL regularization coefficient & $0.02$ & \ref{Sec.3.4} \\
$|\mathcal{G}|$ & Trajectories per group (C / R) & $8$ / $4$ & \ref{Sec.3.4} \\
$N_q$ & Questions per video for scoring & $4$ & \ref{Sec.3.4} \\
\SetCell[c=4]{l,font=\small\bfseries\itshape} BEF and CEF scheduling & & & \\
$s^0$ & Initial construction share & $0.5$ & \ref{Sec.3.5} \\
$s_{\min},s_{\max}$ & Construction-share target bounds & $0.25$, $0.75$ & \ref{Sec.3.5} \\
$\delta_s$ & Maximum share change per cycle & $0.15$ & \ref{Sec.3.5} \\
$\alpha$ & EMA coefficient & $0.5$ & \ref{Sec.3.5} \\
$\rho$ & Uniform-coverage mixture weight & $0.3$ & \ref{Sec.3.5} \\
$K$ & Number of capability categories & $9$ & \ref{Sec.3.5} \\
-- & Minimum support per capability & $10$ probe videos & \ref{Sec.3.5} \\
-- & Lower / upper weight-rate factors & $0.5$ / $2$ & \ref{Sec.3.5} \\
-- & Frontier / exploration / retention & $0.6$ / $0.2$ / $0.2$ & \ref{Sec.3.5} \\
\SetCell[c=4]{l,font=\small\bfseries\itshape} Observation and memory budgets & & & \\
$b_F(T)$ & Supplemental frames per video & $32/64/96/128$ & \ref{Sec.3.2} \\
$b_T$ & Accepted description tokens per video & $4{,}096$ & \ref{Sec.3.2} \\
$B_N$ & Exploration slots / tool-call cap & $12$ & \ref{Sec.3.3} \\
$B_F^R$ & New raw-video frames per question & $64$ & \ref{Sec.3.3} \\
\end{longtblr}

The SFT configuration is summarized in Table~\ref{tab:keyhyperpara}. Both stages use LLaMA-Factory with DeepSpeed ZeRO-3, BF16, and gradient checkpointing. The visual encoder is frozen while the language model and multimodal merger are updated. Table~\ref{tab:app_rl} lists our alternating-RL settings, using \texttt{verl}~\citep{sheng2025hybridflow}, FSDP, and vLLM~\citep{vllm}. Sampling and likelihood recomputation retain the same xgrammar constraints, and the Writer and embedding service remain frozen.

\begin{longtblr}[
caption={Two-stage SFT data and optimization settings. Batch sizes count decision samples; the two stages use different context limits and attention backends.},
label={tab:keyhyperpara}
]{width=\linewidth,colspec={X[l] Q[c,wd=3.0cm] Q[c,wd=3.0cm]},colsep=4pt,rowsep=1.5pt,cells={font=\small},rowhead=1,row{1}={font=\small\bfseries},hline{1,Z}={0.8pt},hline{2}={0.4pt}}
Setting & Stage I & Stage II \\
Videos & 4,205 & 9,995 \\
Complete trajectories & 4,317 & 102,732 \\
Decision samples & 18,932 & 349,717 \\
Epochs & 1 & 1 \\
Peak learning rate & $10^{-5}$ & $10^{-5}$ \\
Effective batch size & 24 & 32 \\
Devices $\times$ per-device batch & $8\times1$ & $8\times1$ \\
Gradient accumulation & 3 & 4 \\
Context limit (tokens) & 65,536 & 32,768 \\
Attention backend & SDPA & FlashAttention-2 \\
Learning-rate schedule & Cosine & Cosine \\
Warm-up fraction & 0.05 & 0.05 \\
Weight decay / gradient clipping & 0 / 1 & 0 / 1 \\
Sequence packing & Off & Off \\
Internal validation split & None & None \\
\end{longtblr}

\begin{longtblr}[
caption={Alternating-RL hyperparameters and runtime settings. Construction observations are budgeted per video; retrieval interactions and raw-video observations are budgeted per question.},
label={tab:app_rl}
]{width=\linewidth,colspec={X[l] Q[l,wd=5.0cm]},colsep=5pt,rowsep=1.5pt,cells={font=\small},rowhead=1,row{1}={font=\small\bfseries},hline{1,Z}={0.8pt},hline{2}={0.4pt}}
Setting & Value \\
Construction / Retrieval learning rate & $2\times10^{-6}$ / $10^{-6}$ \\
Candidates per construction / Retrieval group & 8 / 4 \\
Policy clipping / KL coefficient & 0.2 / 0.02 \\
Regression / invalid-terminal penalty & 1 / 0.2 \\
Efficiency tie-breaking coefficient & 0.05 \\
Rollout / Writer temperature & 0.7 / 0 \\
PPO epochs / entropy coefficient & 1 / 0 \\
Maximum gradient norm & 1 \\
Controller context / output limit (tokens) & 48,000 / 512 \\
Supplemental Writer context / output limit & 12,000 / 2,048 tokens \\
Retrieval rollout context / output limit & 65,536 / 8,192 tokens \\
Supplemental description / step limit & 4,096 / 256 tokens \\
Supplemental frames / observation & 2, 4, 8, or 16 \\
Exploration / final-only decision slots & 12 / 1 \\
Raw-video frames / per-call limit & 64 / 16 \\
Frame maximum long edge & 896 pixels \\
\end{longtblr}

The limits in Table~\ref{tab:app_rl} distinguish model output allowances from accepted memory text. A Writer may generate up to 2,048 tokens for structured output while at most 256 description tokens are accepted at one step. During RL, an invalid Controller decision is recorded without hidden policy resampling. The standalone build mode permits one explicit correction attempt per stage with validation feedback; Writer format repair and bounded transport retries are separate mechanisms. None of these retries constitutes evidence verification or a new learned component.

\subsection{Condensed Prompt Templates}
\label{app:prompts}
The following templates summarize the fixed behavioral instructions; runtime memory excerpts, legal actions, budgets, and JSON schemas are inserted separately. They are condensed descriptions rather than verbatim copies of complete service requests. The Controller learns choices within this interface, not unrestricted prompt rewriting.

\begin{tcolorbox}[breakable,colback=gray!4,colframe=gray!45,coltitle=black,title={Controller: global selection},fonttitle=\bfseries,fontupper=\small]
Select the next Macro for question-blind supplemental memory. Use the Base and Delta summaries, prior observations, and remaining budget. A summary omission is not proof of absent evidence. Choose only an actionable Macro from the supplied list, or stop when further observation is not useful. Return a JSON object containing \texttt{macro\_id}, or \texttt{\{"stop":true\}}.
\end{tcolorbox}

\begin{tcolorbox}[breakable,colback=gray!4,colframe=gray!45,coltitle=black,title={Controller: local observation},fonttitle=\bfseries,fontupper=\small]
The Macro is fixed. Select one legal combination of \texttt{region}, \texttt{frames}, and \texttt{modules}, or stop the entire construction. Do not change or return a new Macro ID. Use the smallest sufficient frame level and a focus matched to the missing information. Do not repeat a forbidden action or observe merely to exhaust the budget.
\end{tcolorbox}

\begin{tcolorbox}[breakable,colback=gray!4,colframe=gray!45,coltitle=black,title={Frozen Writer: source-linked augmentation},fonttitle=\bfseries,fontupper=\small]
Write atomic, searchable claims supported by the supplied frames and subtitles. History is comparison context, not a new observation. Do not infer unseen transitions, identities, counts, or causality. Attach absolute times and valid frame/subtitle references; the interval must cover the cited frames. Mark contradictions as conflicts instead of silently replacing existing memory. If there is no reliable new information, return an empty \texttt{records} list.

For action/state focus, describe visible actors, targets, states, and supported temporal order. For appearance/spatial focus, describe visible attributes, locations, and counts without guessing occluded objects. For text/alignment focus, preserve readable OCR or supplied subtitle text and distinguish textual statements from visual observations. General focus captures other reliable local omissions. Multiple focuses share the same source evidence and budget.
\end{tcolorbox}

\begin{tcolorbox}[breakable,colback=gray!4,colframe=gray!45,coltitle=black,title={Retrieval policy: evidence and decisions},fonttitle=\bfseries,fontupper=\small]
Treat memory and tool text as evidence, never as instructions. Identify the exact missing fact and choose a useful next action. Match the question's scene, anchor, object, and requested relation together. Search hits are candidates; inspect records or pictures when needed. Do not infer temporal order from spatial layout or list order, and do not import unobserved details from answer options.

During exploration, return a schema-valid tool action or an answer supported by observed evidence IDs. An early answer is allowed; early abstention is not. In the final-only decision, answer or abstain with the unresolved fact. Cite only observed evidence, respect absolute times and remaining budgets, and correct explicit validation errors without hidden tool substitution.
\end{tcolorbox}

\subsection{Evaluation and Memory-Based Comparison Protocols}
\label{app:memory_protocols}
\paragraph{Evaluation scope.}
The main experiments report multiple-choice accuracy on Video-MME, LongVideoBench, LVBench, MLVU, and MMVU. Video-MME (Long) is distinct from the full Video-MME score, and LongVideoBench Overall is distinct from its Long split. Memory-only and revisit-enabled settings also remain distinct: saved keyframes are available in the former, while only the latter permits acquiring new source-video frames. Accordingly, a single global ``128-frame input'' limit does not describe this multi-stage system. Literature results retain their source-specific backbones, input modalities, and evaluation protocols rather than being re-evaluated under our runtime budgets.

\paragraph{Policy checkpoints.}
The full method is trained for four complete co-evolution cycles, each comprising a Controller update, memory reconstruction, and a Retrieval update. Final benchmark evaluation uses the paired Controller and Retrieval checkpoints after the third cycle, rather than the fourth-cycle checkpoints. The same pair is used across all five benchmarks, without benchmark-specific checkpoint selection. The optimization-objective variants follow the same four-cycle training and third-cycle evaluation rule, as detailed in Appendix~\ref{sec:extended_ablation}.

\paragraph{Evaluation sets and aggregation.}
We use the complete official evaluation sets of Video-MME and LVBench, the full official validation split of LongVideoBench, all multiple-choice tasks in the MLVU development split, and the multiple-choice portion of the MMVU test split. The Video-MME and LongVideoBench Long results follow the official duration-based subsets. MLVU is reported as the unweighted mean of the individual task accuracies, rather than accuracy pooled over all questions. The core, design, and cold-start ablations use the same evaluation sets and aggregation rules as the main experiments, without separately sampled question subsets.

\paragraph{Evaluation units and reported columns.}
Memory construction operates per video, whereas retrieval and answering operate per question. For the training-based model, the learned Controller constructs a question-independent Delta from the video's fixed Base; the resulting memory and its index are reused across that video's questions. Each question is then answered independently, without policy updates at test time. In Table~\ref{tab:main_exp}, the two Video-MME columns report overall accuracy without and with subtitles, respectively. The overall Video-MME columns in Tables~\ref{tab:ablation} and~\ref{tab:ablation_cold_start} use the with-subtitles setting. Their Long columns, and the Long-subset results in Table~\ref{tab:ablation_design}, are separate metrics rather than alternative names for the overall scores. Table~\ref{tab:ablation_full} adds LongVideoBench Overall and LVBench results for the corresponding design variants.

\paragraph{Subtitle conditions.}
For the with-subtitles Video-MME setting and for LongVideoBench, LVBench, MLVU, and MMVU, official dataset subtitles are supplied to the Base Memory Writer and, together with the selected region's frames, to the frozen Delta Writer. Raw subtitle text is not exposed directly or through retrieval tools to the Retrieval policy, which uses the resulting Writer-generated memory. In the without-subtitles Video-MME setting, no external subtitles or ASR text are supplied during Base construction, Delta augmentation, or question answering; text visible in the video frames remains available. The Video-MME Long columns in the core, design, and cold-start ablations use the with-subtitles protocol. The other four benchmarks retain the same subtitle-input path across the main experiments and ablations.

\paragraph{Benchmark decoding and limits.}
For the training-based model and its ablations, Retrieval uses greedy decoding with temperature zero at benchmark evaluation. Controller decisions for final memory construction and Writer calls for Base extraction, higher-level aggregation, and Delta augmentation also use greedy decoding. The Controller, supplemental Writer, and Retrieval retain context/output limits of 48,000/512, 12,000/2,048, and 65,536/8,192 tokens, respectively (Table~\ref{tab:app_rl}). These settings concern final evaluation, not sampled training trajectories or diagnostic rollouts.

\paragraph{Repeated evaluation.}
For the training-based model and its ablations, we run question answering four times with seeds $\{17,29,43,71\}$ and report the arithmetic mean of the four accuracies, using the benchmark-specific aggregation rules above. Each configuration's memory and index remain fixed across the four runs; neither Base nor Delta is reconstructed between them. The repetitions concern inference with fixed checkpoints and memory, not independent training runs or independently sampled memory constructions. Greedy decoding is retained in all four runs; the seed settings do not imply stochastic answer sampling.

\paragraph{Answer scoring and failure states.}
The implemented multiple-choice scorer normalizes the prediction and reference to option labels and checks exact equality; it does not invoke an LLM judge or grade the reasoning text. Reference answers are supplied only to the scorer after the policy finishes. With a valid reference label, abstentions and invalid terminal decisions that yield no answer receive zero correctness, whereas an invalid intermediate action can still be followed by a correct final answer. Technical execution failures are recorded separately as unscored outcomes, not conflated with policy abstention or invalid termination. Runtime summaries retain scored and unscored counts separately; final benchmark reporting follows the completion protocol below rather than dropping technical failures from the denominator.

\paragraph{Completion and retry protocol.}
For the training-based model and its ablations, only technical execution failures are rerun, retaining the affected run's original seed, checkpoints, memory, and budgets. Completed outcomes are preserved: incorrect answers, abstentions, and invalid policy terminations are not retried. All technical failures are resolved before final aggregation, so each of the four runs covers the full specified evaluation set. Accuracy uses the complete set for the corresponding benchmark, subset, or MLVU task; abstentions and invalid terminations without a valid answer count as incorrect.

\paragraph{Evidence access and stopping.}
The full model permits raw-video revisiting on all five benchmarks, including both Video-MME subtitle conditions. Memory-only is not text-only: the policy can inspect stored keyframes as well as memory records. Revisit-enabled inference additionally permits new source-video observations through \texttt{view\_video}; the policy may still finish without using this tool. The question-time allowance is 64 new frames in total, with at most 16 per call, and is separate from both Base sampling and the duration-dependent augmentation budget in Eq.~\ref{eq:app_frame_budget}. The interaction protocol in Appendix~\ref{app:tools} provides 12 exploration slots and, if needed, a final-only decision. Invalid intermediate actions consume a slot without executing a tool. Consequently, the revisiting ablation concerns access to new source frames, not removal of stored visual evidence or changes to the Writer's parameters.

\paragraph{Training diagnosis versus benchmark inference.}
The 512-question paired probe in Appendix~\ref{app:feedback} is used to schedule subsequent training, not to replace any of the five benchmark evaluations. The probe uses temperature 0.7; final benchmark evaluation instead uses the greedy protocol above, despite sharing the four seed values. BEF and CEF are inactive during benchmark inference: their effect is carried by the learned policies, rather than by further optimization on evaluation questions.

\paragraph{Baseline organization.}
We distinguish query-time video access from explicit memory-based reasoning. Table~\ref{tab:main_exp} covers proprietary models alongside open-source reasoning-enhanced and video-revisiting/tool-based baselines, including their reported results on MMVU~\citep{zhao2025mmvu}. Memory-based methods may also revisit video evidence; tool use alone does not determine the category.

In Table~\ref{tab:main_exp}, ``Baseline'' denotes Qwen3-VL-8B used within our tool-enabled framework without task-specific training of the model. We report our own evaluation for this row. It uses Base memory only, with the same Qwen3.8-27B Writer and Base construction procedure as the training-free variant, and does not run a memory-augmentation Controller or construct a Delta. Neither deployed model undergoes task-specific parameter updates in this study. Across the five benchmarks, the baseline follows the training-based model's evaluation sets, subtitle inputs for Base construction, retrieval tools, and budgets, including raw-video revisiting. It also uses greedy decoding, a Retrieval context/output limit of 65,536/8,192 tokens, and the same four-run accuracy averaging with fixed Base and index, seeds $\{17,29,43,71\}$, and technical-failure-only completion protocol.

Among the literature baselines in Table~\ref{tab:main_exp}, results follow
ParaVT~\citep{yang2026paravt} except for VideoZoomer's original Video-MME
overall (65.2), LongVideoBench validation (57.7), LVBench (41.5), and
MLVU dev (68.8) results~\citep{ding2026videozoomer}. These original results use
up to 128 frames, whereas ParaVT reports 64-frame re-evaluations. VideoZoomer's
65.2 is listed under w/o subtitles, with ``--'' under w/ subtitles.
Its MMVU entry retains ParaVT's result, since the original paper
reports VideoMMLU and VideoMMMU, not MMVU. This row therefore combines
source-specific evaluations.

Table~\ref{tab:memory_exp} lists seven prior methods and both VideoEvolve variants, grouped into training-free and training-based approaches. Following the terminology in the main text, Memory Construction lists the primary memory-writing model or module, and Retrieval lists the main retrieval/answering model; neither column enumerates every auxiliary component. Training-free refers to a memory and retrieval procedure without additional parameter optimization; it does not imply that all component models are untrained. Reusable memory is constructed without the downstream question, whereas query-conditioned memory is updated for each question. MemVid's KV-cache memorizer is a latent-memory module rather than a caption writer.

EgoRAG~\citep{yang2025egolife} builds hierarchical textual memory from clip captions and higher-level summaries. Its original EgoLife experiments vary the captioner, use GPT-4o for final responses in the main evaluation, and also study EgoGPT-based answering. Although EgoGPT is fine-tuned, EgoRAG's hierarchical retrieval procedure itself does not introduce parameter optimization; this is the basis for its grouping in our table. The original paper does not report LVBench or Video-MME (Long) accuracy. We therefore use WorldMM's results of 32.2 and 41.1, respectively. WorldMM identifies a proprietary backbone but does not explicitly specify the captioner/answerer pairing for this evaluation, so both model entries are shown as ``--.''

HippoMM~\citep{lin2026hippomm} segments audiovisual events, consolidates their multimodal representations, and switches between summary retrieval and detailed recall without method-specific gradient training. Its original implementation uses Qwen2.5-VL for visual descriptions and memory summaries, alongside ImageBind embeddings and Whisper transcripts. GPT-4o handles query classification; the revised paper also identifies it as the final answer synthesizer, whereas the initial appendix allows Qwen2.5-VL synthesis with GPT-4o for complex cases. Our model columns summarize these principal components, not a verified checkpoint pairing for WorldMM's evaluation. The original paper does not report either benchmark in our table; 38.2 on LVBench and 41.6 on Video-MME (Long) are taken from WorldMM, which identifies a GPT backbone but does not specify its version.

For M3-Agent~\citep{long2026seeing}, we use the original-paper result of 61.8 on Video-MME (Long), with a Qwen3-32B control policy and a Qwen2.5-Omni-7B memorization model. Memorization is trained with SFT and control with RL in separate stages. The memorization training data combine episodic annotations from Gemini 1.5 Pro and GPT-4o; these are data-synthesis teachers, not the deployed memory model. LVBench accuracy is not reported in this configuration.

MemVid~\citep{yuan2025memory} trains its memorizer to generate memory-grounded retrieval clues through SFT followed by DPO using downstream generation feedback; its retriever and answer generator remain fixed. The reported configuration uses Qwen2-VL-7B as the generator and LanguageBind-Large for retrieval over 10-second moments, with 128 input frames for answering. From its original Tables 1 and 2, we retain LVBench 44.4 and only the with-subtitles Video-MME (Long) result of 57.1. KV-cache describes the memory representation, not a model name. The memorizer's separate backbone is not explicitly identified, so its Memory Construction entry is ``--'' rather than an assumed QA backbone.

Ego-R1~\citep{tian2026ego} trains a Qwen2.5-3B-Instruct controller with SFT and multi-turn RL over a preconstructed hierarchical memory; it does not jointly train memory construction. The original report gives 64.9 on Video-MME (Long). Its benchmark is labeled ``w/o subtitle,'' but Appendix B.3 describes Gemini 1.5 Pro captioning 30-second clips with visual content and transcribed dialogue, followed by GPT-4.1 aggregation into 10-minute summaries. This is therefore not a visual-only pipeline. It also uses proprietary visual tools and a separate summarization model. For compactness, the table lists only Gemini 1.5 Pro as the primary memory writer. The original report does not provide LVBench accuracy, so we supplement it with WorldMM's 34.1. This evaluation uses the released controller checkpoint, changes the hierarchical memory timescales to 10 seconds, 30 seconds, 3 minutes, and 10 minutes, and caps reasoning at five iterations. Thus, the LVBench and Video-MME (Long) entries are from different evaluation configurations; the model columns summarize the original method rather than certify an identical pipeline for both scores.

WorldMM-GPT~\citep{yeo2026worldmm} uses GPT-5-mini to construct episodic and semantic memories, and GPT-5 for retrieval and answering. Memory captioning receives video frames and ASR transcripts. Its LVBench evaluation excludes 15 questions without temporal segment annotations, leaving 1,534 questions; its Video-MME (Long) evaluation contains 900 questions. These question sets also apply to the EgoRAG and HippoMM results taken from WorldMM, and the same LVBench subset applies to Ego-R1's supplemented score. WorldMM aligns the hierarchical memory timescales of EgoRAG and Ego-R1 with its own, and otherwise follows official baseline specifications. Its retrieval agent and iterative baselines are limited to five iterations.

MERIT~\citep{choi2026keep} uses GPT-5-mini to caption 30-second clips and extract multiple episodic keys; the reported 71.8/77.7 results use GPT-5 as the QA solver. It incorporates ASR transcripts for Video-MME (Long), but uses visual frames only and excludes the dialogue key for LVBench. We retain each paper's reported benchmark setting rather than assuming a common input modality or question subset.

The training-based VideoEvolve variant uses Qwen3.5-4B to select memory augmentations, a frozen Qwen3.8-27B Writer, and Qwen3-VL-8B for retrieval and answering. Its Memory Construction entry denotes the Writer, not the learnable Memory Evolver. Table~\ref{tab:memory_exp} reports 58.9 on LVBench and 67.1 on Video-MME (Long) for this variant; the latter is not the full Video-MME score with subtitles. The training-free variant uses Qwen3.8-27B for both memory writing and retrieval, reaching 76.7 and 78.8 on the respective benchmarks. These are separate configurations, not a matched-backbone training ablation. Since frame budgets, transcript access, and question subsets have not been aligned across the listed systems, score differences provide descriptive cross-system comparisons rather than controlled measures of training or architectural gains.

\paragraph{Training-free configuration.}
The training-free VideoEvolve variant uses Base memory only: it does not run a memory-augmentation Controller or construct a Delta. Qwen3.8-27B serves as both the Base Writer and the tool-using Retrieval policy, without SFT, RL, or other task-specific parameter updates in this study. Base sampling and extraction follow Appendix~\ref{app:base_memory}, and question answering retains the stored-keyframe access, raw-video revisiting, and interaction budgets above. The LVBench and with-subtitles Video-MME Long evaluations use the same complete sets and subtitle inputs for Base construction as the training-based variant. Writer and Retrieval decoding is greedy, with a Retrieval context/output limit of 65,536/8,192 tokens. The fixed Base and index are reused for four question-answering runs with seeds $\{17,29,43,71\}$; we average their accuracies and apply the same technical-failure-only retry and full-coverage protocol.

\subsection{Extended Ablation Results}
\label{sec:extended_ablation}

\paragraph{Ablation axes and reference rows.}
The supplementary experiments separate three questions: which observations to add to memory, which objective to use for policy optimization, and how to initialize the Retrieval Evolver before co-evolution. Tables~\ref{tab:ablation_design} and~\ref{tab:ablation_full} organize the first two questions into matching memory-construction and optimization-objective groups; Table~\ref{tab:ablation_cold_start} addresses the third. ``Learned augmentation'' and ``Quality-first (ours)'' report the full-model reference for the construction and objective ablations, respectively. Table~\ref{tab:ablation_design} uses the corresponding Long-subset scores in Table~\ref{tab:ablation}, while Table~\ref{tab:ablation_full} reports the matching LongVideoBench Overall and LVBench scores. The two groups examine different aspects of the method; co-evolution and feedback ablations are reported separately in Table~\ref{tab:ablation}.

\paragraph{Separating training and inference interventions.}
``Memory evolution only'' and ``Retrieval evolution only'' in Table~\ref{tab:ablation} identify which policy is optimized, not which evidence is visible at inference. The feedback-free co-evolution comparison disables both BEF and CEF for joint and single-Evolver training, as described in the main text. By contrast, ``w/o video revisiting'' changes question-time access to new source frames. It is distinct from ``Base memory only'' in the construction group, which omits the Delta but retains the Base records and saved keyframes. Thus, memory evolution, memory augmentation, and memory-only inference denote different interventions.

\paragraph{Core ablation configurations.}
In Table~\ref{tab:ablation}, ``Base (no evolution)'' pairs Base-plus-Delta memory from the initial Controller with the Retrieval policy after both SFT stages; neither policy has undergone co-evolution. ``Memory evolution only'' freezes this post-SFT Retrieval policy, while ``Retrieval evolution only'' keeps the initial Controller and its Base-plus-Delta memory fixed. The two single-policy variants and the four evolution-guidance variants start from the full method's initializations, update their active policies for four rounds, and use the third-round checkpoints for evaluation; inactive policies remain at their initial checkpoints.

For ``w/o alternating updates'', all Controller training is completed with Retrieval frozen; its resulting memory is then fixed throughout the subsequent Retrieval training stage. The initializations and cumulative per-policy update budgets match the full method's four-round training, but evaluation uses the endpoints of the two sequential stages. BEF is disabled and CEF is retained. This variant therefore changes the update ordering and removes dynamic cross-policy allocation while retaining capability-guided sampling.

For ``w/o BEF'', the Memory and Retrieval allocation shares remain fixed at 50\% each throughout training; this does not require equal numbers of groups on the two sides. For ``w/o CEF'', sampling is uniform over all training videos on the Memory side and all training questions on the Retrieval side, without capability-demand or frontier weighting. In contrast, ``w/o frontier diagnosis'' retains capability-demand weights but samples uniformly within the selected capability, removing the frontier/exploration/retention partition.

All of these core variants permit raw-video revisiting at evaluation, including Base (no evolution) and the sequential-training variant. The separate ``w/o video revisiting'' row instead reuses the full model's third-round policy pair and the same memory without retraining, disabling only \texttt{view\_video}. Textual memory and saved keyframes remain accessible.

\textbf{Memory construction.}
In the construction groups of Tables~\ref{tab:ablation_design} and~\ref{tab:ablation_full}, all four strategies are evaluated with the same fixed Retrieval checkpoint used by the full model in the main experiments; Retrieval is not retrained for each memory variant. Learned augmentation also uses the full model's co-evolved Memory Controller checkpoint. All four variants permit raw-video revisiting at question time, including Base memory only, which removes the Delta rather than disabling \texttt{view\_video}.

Table~\ref{tab:ablation_memory} compares augmentation strategies under the same
observation budget within this ablation group. Learned augmentation achieves
the highest accuracy on both Long subsets. Random and uniform augmentation
yield smaller gains, suggesting
that observation selection matters beyond simply adding observations to the
Base Memory.

The learned strategy selects the Macro, temporal region, frame level, and extraction focus through the interface in Appendix~\ref{app:memory_augmentation}. The uniform and random rows provide alternative observation-selection strategies, whereas Base memory only adds no supplemental records. The shared budget is an allowance for augmentation, not a claim that every strategy consumes the same number of frames or has equal runtime; the learned construction policy can stop before exhausting it. Learning the selection policy also does not mean fine-tuning the Writer, which remains frozen.

Uniform augmentation places observations at evenly spaced locations along the full video timeline. It requests eight frames at each location and uses the same generic description instruction for the Writer. Under normal execution, it continues until the observation budget is exhausted, without content-dependent early stopping.

Random augmentation follows the same evenly spaced placement rule and generic Writer instruction as the uniform baseline. At each step, it samples uniformly from \texttt{STOP} and the budget-feasible frame levels in $\{2,4,8,16\}$, excluding levels that exceed the remaining frame budget before sampling. Selecting \texttt{STOP} terminates construction; otherwise, the sampled frame level is requested at the scheduled location. This baseline therefore randomizes the frame level and stopping decision, rather than observation placement or extraction focus. No random seed is explicitly fixed for construction; the resulting memory is held fixed across the four question-answering runs.

\textbf{Optimization objective.}
For each alternative objective in Tables~\ref{tab:ablation_design} and~\ref{tab:ablation_full}, both the Memory Controller and Retrieval policy are retrained from the same initializations as the full method for four co-evolution rounds. Each variant is evaluated with its paired third-round checkpoints, following the same fixed-round selection rule as the full method. BEF and CEF remain enabled and operate on each run's diagnostic results. All objective variants permit raw-video revisiting during evaluation.

Table~\ref{tab:ablation_objective} compares the quality-first objective with
alternative reward designs on the Long subsets of Video-MME and LongVideoBench.
Removing the regression penalty reduces accuracy by 2.2/3.1 points, and a scalar
quality--efficiency reward also underperforms the full objective. Quality-first
optimization improves over quality-only group-relative optimization by 0.9 and
1.2 points, respectively. These accuracy
results support the objective design, but do not by themselves establish an
efficiency advantage.

The full quality-first objective is specified in Appendix~\ref{app:objective}: task-quality advantages are computed first, and the bounded efficiency preference is applied only among eligible, equal-quality trajectories. It does not subtract a resource penalty from the task reward before group normalization. The objective ablation therefore distinguishes the construction regression term, task-quality learning, and the placement of the efficiency preference. These are training signals: the reported endpoint remains multiple-choice accuracy, not a quality--cost reward or a score for the reasoning text. The underlying resource quantities are supplemental frame inputs and accepted text for construction, and executed tools and newly acquired frames for retrieval; they are not end-to-end latency measurements.

The four alternative objectives are defined as follows:
\begin{enumerate}
    \item \textbf{Without regression penalty.} Only the construction reward's regression coefficient is changed to $\lambda_{\mathrm{reg}}=0$. The invalid-termination penalty ($\lambda_{\mathrm{inv}}=0.2$), group normalization, efficiency term, and efficiency-eligibility conditions remain unchanged. The Retrieval reward definition is unchanged, although both policies are retrained as described above.
    \item \textbf{Quality-only group relative.} Both policies use $\beta=0$, so their final advantages contain only the group-normalized task-quality component. The original quality rewards are retained, including $\lambda_{\mathrm{reg}}=1$ for construction and the invalid-termination penalties.
    \item \textbf{Quality minus efficiency cost.} For each policy $p\in\{C,R\}$, the scalar reward is $\widetilde{R}_i^p=R_i^p-0.05E_p^{(i)}$, using the original quality reward and the corresponding cost in Eq.~\ref{eq:app_costs}. Group normalization in Eq.~\ref{eq:app_adv_stability} is applied to $\widetilde{R}_i^p$, not to $R_i^p$. No additional equal-quality efficiency term is applied. The original regression and invalid-termination penalties remain in $R_i^p$.
    \item \textbf{Answer accuracy only.} The construction reward is the mean answer accuracy obtained with the candidate memory, without subtracting reference-memory accuracy. The Retrieval reward is binary final-answer correctness. Both rewards are group-normalized using Eq.~\ref{eq:app_adv_stability}, without regression penalties, separate invalid-termination penalties, or efficiency terms.
\end{enumerate}

\begin{table}[H]
\centering
\caption{\textbf{Memory construction and optimization ablations.}
Accuracy (\%, $\uparrow$) on the Long subsets of Video-MME (VMME) and LongVideoBench (LVB).}
\label{tab:ablation_design}
\begingroup
\renewcommand{\thesubtable}{(\alph{subtable})}
\captionsetup[subtable]{labelformat=simple,skip=3pt}
\begin{subtable}[t]{0.485\linewidth}
\centering
\caption{Memory construction.}
\label{tab:ablation_memory}
\small
\begin{tblr}{
  width=\linewidth,
  colspec={X[l] Q[c,wd=1.1cm] Q[c,wd=0.8cm]},
  colsep=2pt, rowsep=2pt, cells={valign=m},
  row{1}={font=\bfseries}, cell{2}{2,3}={font=\bfseries},
  hline{1,Z}={0.8pt}, hline{2,3}={0.4pt}
}
Augmentation & VMME & LVB \\
Learned augmentation & 67.1 & 65.5 \\
Uniform augmentation & 62.8 & 60.1 \\
Random augmentation & 62.0 & 59.2 \\
Base memory only & 61.0 & 58.7 \\
\end{tblr}
\end{subtable}\hfill
\begin{subtable}[t]{0.485\linewidth}
\centering
\caption{Optimization objective.}
\label{tab:ablation_objective}
\small
\begin{tblr}{
  width=\linewidth,
  colspec={X[l] Q[c,wd=1.1cm] Q[c,wd=0.8cm]},
  colsep=2pt, rowsep=2pt, cells={valign=m},
  row{1}={font=\bfseries}, cell{2}{2,3}={font=\bfseries},
  hline{1,Z}={0.8pt}, hline{2,3}={0.4pt}
}
Objective & VMME & LVB \\
Quality-first (ours) & 67.1 & 65.5 \\
w/o regression penalty & 64.9 & 62.4 \\
Quality-only group relative & 66.2 & 64.3 \\
Quality $-$ efficiency cost & 65.4 & 63.1 \\
Answer accuracy only & 63.8 & 61.1 \\
\end{tblr}
\end{subtable}
\endgroup
\end{table}

\textbf{Additional design-ablation metrics.}
Table~\ref{tab:ablation_full} complements the Long-subset results in
Table~\ref{tab:ablation_design} with LongVideoBench Overall and LVBench accuracy
for the memory-construction and optimization-objective variants. Co-evolution
and evolution-guidance results are reported in Table~\ref{tab:ablation}.
Comparisons are made within each ablation group; the LVBench scores here are
not used to infer gains relative to the main-comparison results in
Table~\ref{tab:main_exp}.

\begin{table}[H]
\centering
\caption{\textbf{Additional metrics for design ablations.} Accuracy (\%, $\uparrow$)
on LongVideoBench (LVB) Overall and LVBench, complementing Table~\ref{tab:ablation_design}.
Bold marks the best result within each group.}
\label{tab:ablation_full}
\begingroup
\small
\begin{tblr}{
  width=\linewidth,
  colspec={X[l] Q[c,wd=3.0cm] Q[c,wd=2.0cm]},
  colsep=4pt, rowsep=2.2pt, cells={valign=m},
  row{1}={font=\bfseries}, row{2,7}={font=\bfseries\itshape},
  cell{3,8}{2,3}={font=\bfseries},
  hline{1,Z}={0.8pt}, hline{2,7}={0.4pt}, hline{4,9}={0.3pt}
}
Variant & LVB Overall & LVBench \\
\SetCell[c=3]{l} (a) Memory construction & & \\
Learned augmentation & 70.2 & 58.9 \\
Uniform augmentation & 65.4 & 48.6 \\
Random augmentation & 64.7 & 47.9 \\
Base memory only & 64.1 & 47.5 \\
\SetCell[c=3]{l} (b) Optimization objective & & \\
Quality-first (ours) & 70.2 & 58.9 \\
w/o regression penalty & 67.9 & 50.8 \\
Quality-only group relative & 69.3 & 52.7 \\
Quality $-$ efficiency cost & 68.4 & 51.6 \\
Answer accuracy only & 66.7 & 49.9 \\
\end{tblr}
\endgroup
\end{table}

\textbf{Cold-start initialization.}
This ablation changes the Retrieval Evolver's initialization before Agentic RL, not whether co-evolution is subsequently performed. The full model uses Stage I followed by Stage II; ``No cold-start SFT'' skips both stages; ``w/o Stage-I SFT'' uses Stage II only; and ``w/o Stage-II SFT'' uses Stage I only. All four configurations retain four rounds of subsequent co-evolution with BEF and CEF enabled, and are evaluated with their third-round policy pairs and raw-video revisiting enabled. The Memory Evolver does not undergo either cold-start SFT stage, and the Writer remains frozen. Thus, the no-SFT row is not the untrained tool-enabled Baseline in Table~\ref{tab:main_exp}: it has still undergone the subsequent co-evolution procedure.

Two-stage SFT provides a crucial foundation for subsequent co-evolution
(Table~\ref{tab:ablation_cold_start}). Without cold-start SFT, LVBench accuracy falls from
58.9\% to 32.8\% despite retaining co-evolution. Removing Stage I or Stage II
reduces it to 54.2\% or 40.6\%, respectively. The larger loss without Stage II
highlights the contribution of broader visual-trajectory supervision, while the
additional benefit of Stage I supports tool-use initialization. Together, the two
stages provide a stronger starting point than either stage alone.

\begin{table}[H]
\centering
\caption{\textbf{Cold-start initialization of the Retrieval Evolver.}
Overall Video-MME uses the with-subtitles setting.}
\label{tab:ablation_cold_start}
\begingroup
\small
\begin{tblr}{
  width=\linewidth,
  colspec={X[l] Q[c,wd=1.15cm] Q[c,wd=1.0cm] Q[c,wd=1.27cm] Q[c,wd=1.2cm] Q[c,wd=1.35cm] Q[c,wd=1.05cm] Q[c,wd=1.15cm]},
  colsep=2.5pt, rowsep=1.3pt, cells={valign=m},
  row{1}={font=\bfseries}, row{2}={font=\itshape},
  row{3}={bg=blue!6}, cell{3}{2-8}={font=\bfseries},
  hline{1,Z}={0.8pt}, hline{3,4}={0.4pt},
  hline{2}={2-3,4-5}{0.3pt}
}
\SetCell[r=2]{l} Model & \SetCell[c=2]{c} Video-MME & & \SetCell[c=2]{c} LongVideoBench & & \SetCell[r=2]{c} LVBench & \SetCell[r=2]{c} MLVU & \SetCell[r=2]{c} MMVU \\
& overall & long & overall & long & & & \\
VideoEvolve & 73.9 & 67.1 & 70.2 & 65.5 & 58.9 & 72.4 & 75.1 \\
No cold-start SFT & 54.4 & 48.2 & 45.9 & 45.4 & 32.8 & 57.1 & 62.5 \\
w/o Stage-I SFT & 70.2 & 63.5 & 67.7 & 62.5 & 54.2 & 69.9 & 70.1 \\
w/o Stage-II SFT & 64.5 & 57.9 & 62.1 & 51.9 & 40.6 & 60.3 & 65.7 \\
\end{tblr}
\endgroup
\end{table}

\subsection{Limitations and Future Work}
\label{app:limitations}

Although VideoEvolve improves long-video understanding through memory--retrieval co-evolution, several limitations remain. First, the reported evaluation focuses on offline multiple-choice video understanding; it does not establish effectiveness for streaming video, open-ended dialogue, or tasks requiring direct audio understanding. Second, the framework depends on the perceptual quality of the frozen Writer and the evidence available within finite observation budgets. Learned augmentation and selective revisiting may still miss brief events or subtle visual details, while source-linked records and valid citations do not guarantee factual correctness. Third, policy optimization relies on ground-truth answers, and BEF/CEF scheduling uses a fixed diagnostic pool and a predefined capability taxonomy. These signals cover only the questions and capabilities represented in the available data; bottleneck estimates are operational proxies rather than causal attributions of failure. Finally, alternating training requires candidate rollouts, frozen-policy scoring, and repeated memory and index reconstruction. At inference, reusable memory amortizes construction across questions, but multi-step retrieval and optional video revisiting still incur computation. The resource proxies used in our objective do not directly measure end-to-end latency or hardware cost.

Future work can extend VideoEvolve along these directions. Incremental memory updates for streaming inputs, together with richer audiovisual evidence and open-ended evaluation, would broaden the settings in which co-evolution can be studied. Evidence-level verification and calibrated uncertainty could help identify unreliable memory records and determine when additional observation or abstention is appropriate. Another direction is to combine weaker supervision with adaptive capability discovery, reducing dependence on fixed labeled pools while explicitly validating the reliability of the resulting feedback. More efficient memory and index updates, policy distillation, and cost-aware execution could reduce training and deployment overhead. Such extensions should be assessed with end-to-end measurements that separate preprocessing, training, and question-time costs, alongside broader evaluation across video durations, domains, and backbone models.

\par\clearpage
\endgroup

\end{document}